\documentclass[10pt]{article} 
\usepackage[preprint]{tmlr}

\usepackage{amsmath,amsfonts,bm}

\def\eqref#1{equation~\ref{#1}}

\def\floor#1{\lfloor #1 \rfloor}
\def\1{\bm{1}}

\DeclareMathAlphabet{\mathsfit}{\encodingdefault}{\sfdefault}{m}{sl}
\SetMathAlphabet{\mathsfit}{bold}{\encodingdefault}{\sfdefault}{bx}{n}

\usepackage{hyperref}
\usepackage{url}

\usepackage{amsmath,amsthm,amssymb}

\usepackage{mathtools}
\usepackage{subcaption}
\usepackage{enumitem}
\usepackage{algorithm}
\usepackage{algpseudocode}
\usepackage{booktabs}
\usepackage{multirow}
\usepackage{bbm}
\usepackage{adjustbox}

\numberwithin{equation}{section}
\newtheorem{theorem}{Theorem}[section]
\newtheorem{proposition}[theorem]{Proposition}

\theoremstyle{definition}

\theoremstyle{remark}
\newtheorem{remark}[theorem]{Remark}

\title{CLAPS: Aleatoric-Epistemic Scaling via Last-Layer Laplace for Conformal Regression}

\author{\name Dongseok Kim\thanks{These authors contributed equally.} \email jkds5920@gachon.ac.kr \\
      \addr Department of Computer Engineering\\
      Gachon University
      \AND
      \name Hyoungsun Choi\footnotemark[1] \email hschoi@gachon.ac.kr \\
      \addr Department of Computer Engineering\\
      Gachon University
      \AND
      \name Mohamed Jismy Aashik Rasool\footnotemark[1] \email aashikrasool@gachon.ac.kr \\
      \addr Department of Computer Engineering\\
      Gachon University
      \AND
      \name Gisung Oh\thanks{Corresponding author.} \email eustia@gachon.ac.kr \\
      \addr Department of Computer Engineering\\
      Gachon University}

\def\month{10}  
\def\year{2026} 
\def\openreview{\url{https://openreview.net/forum?id=RvoPfNBjwp}} 

\begin{document}

\maketitle

\begin{abstract}
Conformal regression provides finite-sample marginal coverage, but interval efficiency depends on how uncertainty is allocated across inputs. We propose \emph{Conformal Laplace-Aware Predictive Scaling} (CLAPS), which combines a learned heteroscedastic scale with last-layer Laplace uncertainty. The learned variance also weights the training features used to construct the posterior, yielding a noise-weighted support correction in the learned representation. The correction contracts as weighted feature information increases and becomes most influential under localized support imbalance. Experiments show near-nominal marginal coverage and modest efficiency improvements over the matched aleatoric baseline on real regression benchmarks.
\end{abstract}

\section{Introduction}
\label{sec:introduction}

Prediction intervals are a standard tool for uncertainty quantification in regression, but their practical value depends on how interval width varies across inputs. Wider intervals may reflect high response variability or limited training support, and these sources of uncertainty can imply different downstream actions such as additional data collection, human review, or more cautious decisions.

Conformal prediction provides finite-sample marginal coverage under exchangeability, but the guarantee does not determine how interval width should be allocated across the input space. Standard residual-based intervals use a global scale, whereas locally adaptive methods use input-dependent scales to account for heterogeneous predictive difficulty.

Existing conformal methods have incorporated heteroscedasticity, model uncertainty, and feature support through different score constructions. CLAPS does not introduce a new conformal calibration rule or coverage guarantee. Instead, it instantiates standard locally adaptive split conformal prediction with a specific heteroscedastic neural-linear scale obtained from a single fitted regressor.

We propose \emph{Conformal Laplace-Aware Predictive Scaling} (CLAPS), which combines a learned input-dependent observation scale with last-layer Laplace posterior uncertainty. The learned variance controls both the test-time observation scale and each training feature's contribution to the last-layer posterior precision, producing a noise-weighted support correction in the learned representation. The contribution of CLAPS lies in the construction and characterization of this scale, its noise-weighted posterior geometry, and the identification of the regimes in which the correction affects interval allocation.

The correction is most relevant when weighted feature support is uneven and contracts as feature information increases, causing CLAPS to approach the matched aleatoric locally adaptive scale in well-supported regimes.

This paper makes the following contributions:
\begin{itemize}
    \item \textbf{Heteroscedastic last-layer conformal scaling.}
    CLAPS combines a learned observation scale with last-layer posterior uncertainty in a split conformal score.

    \item \textbf{Noise-weighted posterior geometry.}
    The learned variance controls each training feature's contribution to posterior precision, producing a support-sensitive correction in the learned representation.

    \item \textbf{A single-model uncertainty construction.}
    Both observation and support components are obtained from one heteroscedastic neural regressor and a post-hoc last-layer covariance.

    \item \textbf{Graceful reduction to aleatoric local scaling.}
    The posterior correction contracts as weighted feature information increases, recovering the matched LACP scale in well-supported regimes.
\end{itemize}

\section{Related Work}
\label{sec:related_work}

\subsection{Conformal Prediction for Regression}
\label{subsec:conformal_regression}

Conformal prediction provides a general framework for constructing prediction sets with finite-sample marginal coverage under exchangeability. Early work introduced transductive and inductive conformal prediction, nonconformity scores, and rank-based calibration as model-agnostic tools for predictive inference~\citep{saunders1999transduction, papadopoulos2002inductive, gammerman2007hedging, shafer2008tutorial, papadopoulos2011regression}. In regression, these principles yield prediction intervals by calibrating residual-based scores on held-out or resampled data.

Regression conformal methods provide different trade-offs among statistical efficiency, computational cost, and data usage. Full conformal inference evaluates candidate labels jointly with the training sample, split conformal inference separates model fitting and calibration, and jackknife or cross-validation variants reuse observations across multiple fits~\citep{johansson2014regression, lei2018distribution, ndiaye2019computing, barber2021predictive, steinberger2023conditional}. These procedures share the same rank-calibration principle despite differences in score construction and data reuse.

Further developments cover conditional quantile models, covariate shift, online prediction, non-exchangeable data, and training-conditional or conditional-coverage analyses~\citep{romano2019conformalized, tibshirani2019conformal, foygel2021limits, gibbs2021adaptive, barber2023conformal, bian2023training, gibbs2024conformal, oliveira2024split}. This literature establishes the nonconformity score as a central design choice for allocating calibrated uncertainty across inputs.

\subsection{Locally Adaptive and Distributional Conformal Methods}
\label{subsec:adaptive_distributional_conformal}

Locally adaptive conformal methods use input-dependent scales or weights to vary interval width with predictive difficulty. Normalized nonconformity scores divide residuals by an estimated local scale, and related approaches use local weighting, learned features, score transformations, and group-conditional calibration~\citep{papadopoulos2008normalized, seedat2023improving, guan2023localized, hore2025conformal, gibbs2025conformal, jung2022batch, bastani2022practical, kiyani2024conformal, kiyani2024length}. Leverage-based scaling further connects interval allocation to design geometry through input-dependent statistical leverage~\citep{fadnavis2026leverage}.

Distributional conformal methods construct scores from estimates of the conditional response distribution. Existing approaches calibrate conditional distributions, histograms, conditional samples, discretized response models, normalizing flows, and shape templates for multimodal regions~\citep{chernozhukov2021distributional, sesia2021conformal, wang2023probabilistic, plassier2024probabilistic, guha2024conformal, colombo2024normalizing, tumu2024multi, plassier2025rectifying, van2024self}. These methods improve local adaptivity by incorporating more information about the conditional response than absolute residuals alone.

Recent methods incorporate epistemic uncertainty directly into conformal score design. UACQR uses uncertainty in fitted conditional quantiles to adapt conformalized quantile intervals, and EPICSCORE provides a general transformation that combines Bayesian epistemic information with a chosen conformal score~\citep{rossellini2024integrating, cabezas2025epistemic}. CLAPS focuses on a heteroscedastic neural regressor and obtains its support correction from a variance-weighted last-layer posterior.

Adaptive score design has also been studied in classification through adaptive prediction sets, regularized conformity scores, conformal training, and label-ranking methods~\citep{romano2020classification, angelopoulos2020uncertainty, stutz2021learning, huang2024conformal}. Across regression and classification, these results show that score construction controls how prediction-set size responds to local predictive difficulty.

\subsection{Bayesian and Epistemic Uncertainty in Prediction Intervals}
\label{subsec:bayesian_epistemic_uncertainty}

Bayesian predictive modeling distinguishes aleatoric uncertainty from uncertainty about the fitted predictor. Bayesian neural networks represent predictive uncertainty through distributions over weights, functions, or predictive laws, commonly using approximate inference procedures suited to modern architectures~\citep{graves2011practical, hernandez2015probabilistic, blundell2015weight, kingma2015variational, gal2016dropout}.

Scalable Bayesian deep learning methods include variational inference, dropout-based approximations, normalizing-flow posteriors, Bayesian hypernetworks, function-space approximations, and modular Bayesian layers~\citep{louizos2017multiplicative, gal2017concrete, krueger2017bayesian, sun2019functional, tran2019bayesian}. Deep ensembles, evidential models, and deterministic uncertainty methods provide alternative estimates without full posterior inference~\citep{lakshminarayanan2017simple, sensoy2018evidential, amini2020deep, van2020uncertainty, charpentier2020posterior}.

Studies of uncertainty calibration and dataset shift examine the reliability of these estimates under model misspecification and changing input distributions~\citep{kendall2017uncertainties, kuleshov2018accurate, ovadia2019can, maddox2019simple, wilson2020bayesian}. The role of aleatoric and epistemic components in conformal prediction has also been analyzed directly~\citep{sale2025aleatoric}. CLAPS contributes a specific neural-linear construction in which last-layer posterior uncertainty guides local interval allocation under standard conformal calibration.

\subsection{Last-Layer Laplace Approximation for Neural Uncertainty}
\label{subsec:last_layer_laplace}

Laplace approximation constructs an approximate Gaussian posterior around a trained neural-network solution using local curvature information. Modern developments provide scalable implementations, post-hoc uncertainty estimates, marginal-likelihood objectives, and function-space prior formulations~\citep{ritter2018scalable, daxberger2021laplace, kristiadi2021learnable, immer2021scalable, cinquin2024fsp}.

Linearized and subnetwork Laplace methods reduce posterior complexity by approximating the network locally or restricting inference to selected parameter directions. These methods connect neural uncertainty to generalized linear models, Gaussian processes, neural tangent kernels, and efficient adaptation procedures~\citep{immer2021improving, antoran2022adapting, daxberger2021bayesian, deng2022accelerated, ortega2023variational, khan2019approximate, jacot2018neural, maddox2021fast}.

Last-layer Bayesian methods fix the learned representation and place a posterior on the final predictive layer, yielding an efficient neural-linear model~\citep{kristiadi2020being, watson2021latent, harrison2024variational, brunzema2024bayesian}. Their posterior geometry depends on the likelihood and curvature approximation, motivating Kronecker-factored, Gauss--Newton, and automatic-differentiation-based curvature methods as well as improved heteroscedastic regression objectives~\citep{martens2015optimizing, botev2017practical, dangel2019backpack, eschenhagen2023kronecker, seitzer2022pitfalls, stirn2023faithful}. CLAPS uses this last-layer structure with heteroscedastic precision weighting, linking the learned observation scale to predictive variance and feature-level information.

\section{Method}
\label{sec:method}

We consider regression with inputs \(X\in\mathcal{X}\), responses \(Y\in\mathbb{R}\), a training set, a calibration set, and a target miscoverage level \(\alpha\in(0,1)\). CLAPS constructs split conformal intervals using a heteroscedastic last-layer Laplace predictive scale.

The base model maps each input \(x\) to a representation \(\phi(x) \in \mathbb{R}^d\). For notational simplicity, the final-layer bias is absorbed into \(\phi(x)\) by appending a constant coordinate, and the predictive mean is
\[
\mu(x) = \phi(x)^\top w.
\]
The model is trained under the heteroscedastic Gaussian likelihood
\[
    Y\mid x,w\sim\mathcal{N}\bigl(\phi(x)^\top w,h^2(x)\bigr),
\]
where the input-dependent variance head is parameterized as
\[
h^2(x) = \operatorname{softplus}(g(x)) + \epsilon,
\]
with variance floor \(\epsilon > 0\). After training, the representation, variance head, and final-layer point estimate are fixed, and a post-hoc Gaussian covariance is constructed for the final-layer weights.

Let \(H\in\mathbb{R}^{n_{\mathrm{tr}}\times d}\) denote the training feature matrix with rows \(\phi(x_i)^\top\), and define
\[
    W=\mathrm{diag}\left(
    \frac{1}{h^2(x_1)},\ldots,
    \frac{1}{h^2(x_{n_{\mathrm{tr}}})}
    \right).
\]
The weighting assigns less posterior information to observations with larger estimated variance and more information to observations with smaller estimated variance. Given prior precision \(\lambda>0\), the last-layer posterior covariance is
\[
    \Sigma=\left(\lambda I+H^\top W H\right)^{-1}.
\]
The implementation adds a small diagonal jitter to the precision matrix before inversion for numerical stability.

CLAPS defines the predictive variance as
\[
    v(x)=h^2(x)+\phi(x)^\top\Sigma\phi(x).
\]
The first term represents the learned observation variance, and the second represents last-layer posterior variance conditional on the fixed representation. Their sum defines a predictive variance with a noise-weighted support correction in the learned feature space.

For each calibration example \((x_i,y_i)\), the normalized score is
\[
    A(x_i,y_i)=
    \frac{|y_i-\mu(x_i)|}{\sqrt{v(x_i)}}.
\]
Let \(q_\alpha\) be the
\(\lceil(n_{\mathrm{cal}}+1)(1-\alpha)\rceil\)-th order statistic of the calibration scores, using the standard conservative convention when the index exceeds \(n_{\mathrm{cal}}\). The prediction interval for a test input \(x\) is
\[
    C_\alpha(x)=
    \left[
    \mu(x)-q_\alpha\sqrt{v(x)},
    \;
    \mu(x)+q_\alpha\sqrt{v(x)}
    \right].
\]

When \(\lambda\) is tuned, the fitted regressor is held fixed and the training set is divided into covariance-fitting and validation subsets. Candidate values are compared by predictive negative log-likelihood on the validation subset, and the final covariance is recomputed from the complete training set using the selected value. Calibration and test responses remain outside this selection procedure.

The complete procedure is summarized in Algorithm~\ref{alg:claps}; the covariance is constructed once after model fitting in \(O(n_{\mathrm{tr}}d^2+d^3)\) time with \(O(d^2)\) storage, and each test input requires feature extraction and an \(O(d^2)\) quadratic-form evaluation.

\begin{algorithm}[ht]
\caption{Conformal Laplace-Aware Predictive Scaling (CLAPS)}
\label{alg:claps}
\begin{algorithmic}[1]
\Require Training set $\mathcal{D}_{\mathrm{tr}}$, calibration set $\mathcal{D}_{\mathrm{cal}}$, target miscoverage level $\alpha$, prior precision $\lambda>0$ or candidate grid $\Lambda$
\Ensure Prediction interval $C_\alpha(x)$ for a test input $x$

\State Fit a heteroscedastic neural regressor on $\mathcal{D}_{\mathrm{tr}}$ to obtain $\phi(x)$, $\mu(x)=\phi(x)^\top\hat{w}$, and $h^2(x)$.

\If{a candidate grid $\Lambda$ is provided}
    \State Select \(\lambda \in \Lambda\) by predictive negative log-likelihood on a training-only inner split, with the fitted regressor held fixed.
\EndIf

\State Form the training feature matrix $H$ and the weight matrix
\[
W=\mathrm{diag}\left(
\frac{1}{h^2(x_1)},\ldots,
\frac{1}{h^2(x_{n_{\mathrm{tr}}})}
\right).
\]

\State Compute
\[
\Sigma=\left(\lambda I+H^\top W H\right)^{-1}.
\]

\For{each $(x_i,y_i)\in\mathcal{D}_{\mathrm{cal}}$}
    \State Compute
    \[
    v(x_i)=h^2(x_i)+\phi(x_i)^\top\Sigma\phi(x_i),
    \qquad
    A_i=\frac{|y_i-\mu(x_i)|}{\sqrt{v(x_i)}}.
    \]
\EndFor

\State Set
\[
k=\left\lceil(n_{\mathrm{cal}}+1)(1-\alpha)\right\rceil
\]
and let $q_\alpha$ be the $k$-th order statistic of the calibration scores, using the standard conservative convention if $k>n_{\mathrm{cal}}$.

\For{a test input $x$}
    \State Compute
    \[
    v(x)=h^2(x)+\phi(x)^\top\Sigma\phi(x).
    \]
    \State Return
    \[
    C_\alpha(x)=
    \left[
    \mu(x)-q_\alpha\sqrt{v(x)},
    \mu(x)+q_\alpha\sqrt{v(x)}
    \right].
    \]
\EndFor
\end{algorithmic}
\end{algorithm}

\section{Theory}
\label{sec:theory}

\subsection{Finite-Sample Validity under Adaptive Scaling}
\label{subsec:validity}

Let \(\mathcal{D}_{\mathrm{tr}}=\{(X_i,Y_i)\}_{i=1}^{n_{\mathrm{tr}}}\) and \(\mathcal{D}_{\mathrm{cal}}=\{(X_i,Y_i)\}_{i=n_{\mathrm{tr}}+1}^{n_{\mathrm{tr}}+n_{\mathrm{cal}}}\), and write \(n=n_{\mathrm{tr}}+n_{\mathrm{cal}}\). After fitting \(\mu(\cdot)\), \(h^2(\cdot)\), \(\phi(\cdot)\), and \(\Sigma\) on \(\mathcal{D}_{\mathrm{tr}}\), define
\[
    v(x)=h^2(x)+\phi(x)^\top\Sigma\phi(x),
    \qquad
    A(x,y)=\frac{|y-\mu(x)|}{\sqrt{v(x)}}.
\]
Assume \(v(x)>0\) on the input domain. For calibration scores \(A_i=A(X_i,Y_i)\), \(i=n_{\mathrm{tr}}+1,\ldots,n\), let \(q_\alpha\) be the \(\lceil(n_{\mathrm{cal}}+1)(1-\alpha)\rceil\)-th order statistic, with \(q_\alpha=\infty\) if this index exceeds \(n_{\mathrm{cal}}\). The resulting interval is
\[
    C_\alpha(x)=
    \left[
    \mu(x)-q_\alpha\sqrt{v(x)},
    \;
    \mu(x)+q_\alpha\sqrt{v(x)}
    \right].
\]

\begin{proposition}[Finite-sample validity]
\label{prop:finite_sample_validity}
Suppose that the calibration examples and the test point \((X_{n+1},Y_{n+1})\) are exchangeable conditional on \(\mathcal{D}_{\mathrm{tr}}\). Then
\[
    \mathbb{P}\left\{Y_{n+1}\in C_\alpha(X_{n+1})\right\}
    \geq
    1-\alpha.
\]
\end{proposition}

\begin{proof}
Condition on \(\mathcal{D}_{\mathrm{tr}}\), so that \(A(\cdot,\cdot)\) is fixed. Exchangeability of the calibration examples and the test point implies exchangeability of
\[
    A_{n_{\mathrm{tr}}+1},\ldots,A_n,A_{n+1},
    \qquad
    A_{n+1}=A(X_{n+1},Y_{n+1}).
\]
Hence the conformal rank argument gives
\[
    \mathbb{P}\left\{A(X_{n+1},Y_{n+1})\leq q_\alpha \mid \mathcal{D}_{\mathrm{tr}}\right\}
    \geq
    1-\alpha.
\]
The event \(A(X_{n+1},Y_{n+1})\leq q_\alpha\) is equivalent to
\[
    |Y_{n+1}-\mu(X_{n+1})|
    \leq
    q_\alpha\sqrt{v(X_{n+1})},
\]
which is \(Y_{n+1}\in C_\alpha(X_{n+1})\). Taking expectation over \(\mathcal{D}_{\mathrm{tr}}\) completes the proof.
\end{proof}

\subsection{Aleatoric--Epistemic Predictive Scale}
\label{subsec:predictive_scale}

After training, CLAPS treats the representation \(\phi(\cdot)\), variance function \(h^2(\cdot)\), and final-layer point estimate \(\widehat{w}\) as fixed, and uses the last-layer Gaussian approximation
\[
    w\mid\mathcal{D}_{\mathrm{tr}}
    \approx
    \mathcal{N}(\widehat{w},\Sigma).
\]
Together with
\[
    Y\mid x,w
    \sim
    \mathcal{N}(\phi(x)^\top w,h^2(x)),
\]
this yields the local predictive scale used in the conformal score.

\begin{proposition}[Aleatoric--epistemic variance decomposition]
\label{prop:variance_decomposition}
Under the fitted representation, heteroscedastic variance function, and last-layer Gaussian posterior above,
\[
    \operatorname{Var}(Y\mid x,\mathcal{D}_{\mathrm{tr}})
    =
    h^2(x)+\phi(x)^\top\Sigma\phi(x).
\]
\end{proposition}

\begin{proof}
The law of total variance gives
\[
    \operatorname{Var}(Y\mid x,\mathcal{D}_{\mathrm{tr}})
    =
    \mathbb{E}\left[\operatorname{Var}(Y\mid x,w,\mathcal{D}_{\mathrm{tr}})\mid x,\mathcal{D}_{\mathrm{tr}}\right]
    +
    \operatorname{Var}\left(\mathbb{E}[Y\mid x,w,\mathcal{D}_{\mathrm{tr}}]\mid x,\mathcal{D}_{\mathrm{tr}}\right).
\]
The first term equals \(h^2(x)\), and the second term equals
\[
    \operatorname{Var}\left(\phi(x)^\top w\mid\mathcal{D}_{\mathrm{tr}}\right)
    =
    \phi(x)^\top\Sigma\phi(x).
\]
Combining the two terms gives the claim.
\end{proof}

The term \(h^2(x)\) captures input-dependent observation noise, while \(\phi(x)^\top\Sigma\phi(x)\) captures final-layer posterior uncertainty under the learned representation. The resulting scale increases in noisy regions through the aleatoric term and in weakly supported feature regions through the epistemic term.

\subsection{Heteroscedastic Last-Layer Laplace Covariance}
\label{subsec:laplace_covariance}

For fixed \(\phi(\cdot)\) and \(h^2(\cdot)\), the final-layer negative log-likelihood is
\[
    \ell(w)
    =
    \frac{1}{2}\sum_{i=1}^{n_{\mathrm{tr}}}
    \frac{\left(y_i-\phi(x_i)^\top w\right)^2}{h^2(x_i)}.
\]
With prior \(w\sim\mathcal{N}(0,\lambda^{-1}I)\), the negative log-posterior is
\[
    \mathcal{L}(w)
    =
    \frac{1}{2}\sum_{i=1}^{n_{\mathrm{tr}}}
    \frac{\left(y_i-\phi(x_i)^\top w\right)^2}{h^2(x_i)}
    +
    \frac{\lambda}{2}\|w\|_2^2.
\]
Let \(H\in\mathbb{R}^{n_{\mathrm{tr}}\times d}\) have rows \(\phi(x_i)^\top\), and define
\[
    W=\mathrm{diag}\left(\frac{1}{h^2(x_1)},\ldots,\frac{1}{h^2(x_{n_{\mathrm{tr}}})}\right).
\]
Then
\[
    \nabla_w^2\mathcal{L}(w)
    =
    H^\top W H+\lambda I,
    \qquad
    \Sigma
    =
    \left(\lambda I+H^\top W H\right)^{-1}.
\]

\begin{proposition}[Heteroscedastic precision weighting]
\label{prop:heteroscedastic_precision}
The last-layer posterior precision satisfies
\[
    \Sigma^{-1}
    =
    \lambda I+\sum_{i=1}^{n_{\mathrm{tr}}}
    \frac{\phi(x_i)\phi(x_i)^\top}{h^2(x_i)}.
\]
\end{proposition}

\begin{proof}
By the definitions of \(H\) and \(W\),
\[
    H^\top W H
    =
    \sum_{i=1}^{n_{\mathrm{tr}}}
    \frac{\phi(x_i)\phi(x_i)^\top}{h^2(x_i)}.
\]
Adding \(\lambda I\) gives the stated precision matrix.
\end{proof}

Thus, \(h^2(x)\) enters CLAPS both as the test-time aleatoric component of \(v(x)\) and through the training weights \(1/h^2(x_i)\) in the last-layer precision. Noisier training examples contribute less curvature, while lower-noise examples contribute more. The epistemic term is therefore shaped jointly by feature geometry and the heteroscedastic noise pattern.

\begin{remark}[Noise-weighted feature geometry]
Let \(z_i=\phi(x_i)/h(x_i)\) and \(z(x)=\phi(x)/h(x)\). Then
\[
\frac{\phi(x)^\top\Sigma\phi(x)}{h^2(x)}
=
z(x)^\top
\left(
\lambda I+\sum_{i=1}^{n_{\mathrm{tr}}} z_i z_i^\top
\right)^{-1}
z(x),
\]
so the relative posterior correction has the form of a noise-weighted ridge-leverage quantity in the learned representation.
\end{remark}

\subsection{Special Cases and Graceful Reduction}
\label{subsec:reduction}

The CLAPS scale
\[
    v_{\mathrm{CLAPS}}(x)=h^2(x)+\phi(x)^\top\Sigma\phi(x)
\]
contains several standard scale choices as limiting cases. A constant scale \(v(x)=\sigma^2\) recovers residual-based split conformal prediction up to rescaling of the conformal quantile. Removing the epistemic term gives
\[
    v_{\mathrm{LACP}}(x)=h^2(x),
\]
the locally adaptive aleatoric scale. Taking the aleatoric variance to be constant while retaining posterior uncertainty gives
\[
    v_{\mathrm{post}}(x)=\sigma^2+\phi(x)^\top\Sigma\phi(x).
\]
CLAPS retains both the input-dependent aleatoric term and the last-layer epistemic term.

Now consider a sequence of training sets with
\[
    \Sigma_n=\left(\lambda I+H_n^\top W_n H_n\right)^{-1},
    \qquad
    v_n(x)=h^2(x)+\phi(x)^\top\Sigma_n\phi(x).
\]

\begin{proposition}[Graceful reduction to aleatoric scaling]
\label{prop:graceful_reduction}
Suppose \(h^2(x)>0\) and
\[
    \phi(x)^\top\Sigma_n\phi(x)\to 0.
\]
Then
\[
    v_n(x)\to h^2(x),
    \qquad
    \frac{\sqrt{v_n(x)}}{h(x)}\to 1.
\]
\end{proposition}

\begin{proof}
Since
\[
    v_n(x)=h^2(x)+\phi(x)^\top\Sigma_n\phi(x),
\]
the first convergence follows immediately. Because \(h^2(x)>0\),
\[
    \frac{\sqrt{v_n(x)}}{h(x)}
    =
    \sqrt{1+\frac{\phi(x)^\top\Sigma_n\phi(x)}{h^2(x)}}
    \to
    1.
\]
\end{proof}

A sufficient condition is
\[
    \lambda_{\min}\left(\lambda I+H_n^\top W_n H_n\right)\to\infty
    \qquad
    \text{and}
    \qquad
    \|\phi(x)\|_2<\infty.
\]
Indeed,
\[
    0\leq \phi(x)^\top\Sigma_n\phi(x)
    \leq
    \frac{\|\phi(x)\|_2^2}{\lambda_{\min}\left(\lambda I+H_n^\top W_n H_n\right)}
    \to 0.
\]
Therefore, as weighted feature information grows, the last-layer epistemic contribution vanishes and the CLAPS scale reduces to the heteroscedastic aleatoric scale.

\section{Experiments}
\label{sec:experiments}

\subsection{Common Experimental Design}
\label{subsec:common_exp_design}

All experiments target marginal coverage \(1-\alpha=0.90\). The synthetic studies use independent training, calibration, and test samples and are repeated over 30 random seeds. The real-data benchmark is repeated over 20 random seeds using a \(60\%/20\%/20\%\) training, calibration, and test split, with at most \(5{,}000\) observations per dataset.

The shared heteroscedastic neural regressor learns a predictive mean \(\mu(x)\), an input-dependent variance \(h^2(x)\), and a representation \(\phi(x)\). Within controlled comparisons, LACP, Unweighted Laplace, LWCP, and CLAPS use the same fitted regressor and representation. LACP uses the normalization scale \(h(x)\). Unweighted Laplace uses
\[
\sqrt{h^2(x)+\phi(x)^\top\Sigma_{\mathrm{unw}}\phi(x)},
\qquad
\Sigma_{\mathrm{unw}}=\left(\lambda I + H^\top H\right)^{-1}.
\]
LWCP uses ridge leverage in the learned representation. CLAPS uses
\[\sqrt{h^2(x)+\phi(x)^\top\Sigma\phi(x)},
\qquad
\Sigma=\left(\lambda I + H^\top W H\right)^{-1},
\qquad
W_{ii}=\frac{1}{h^2(x_i)}.
\]
For Unweighted Laplace and CLAPS, the prior precision is selected by predictive negative log-likelihood on a training-only inner split with the fitted regressor held fixed.

The broader comparisons include Split CP, CV+, CQR, UACQR-S, and EPICSCORE-MDN. UACQR-S uses Monte Carlo dropout variation in the estimated conditional quantiles as an epistemic scaling signal. EPICSCORE-MDN models the conditional distribution of normalized conformity scores with a dropout mixture-density network. The symmetric Gaussian DCP implementation used in the original manuscript coincides algebraically with LACP and is represented by the LACP results.

We report marginal and region-level coverage, interval width, and interval score. Lower interval scores indicate better joint performance in width and coverage. Support-dependent allocation is evaluated through regional width ratios in the synthetic studies and nearest-neighbor support strata in the real-data study. Tables report means and standard errors. Pairwise comparisons use paired bootstrap confidence intervals, with a hierarchical dataset--seed bootstrap for the real-data benchmark.

\subsection{Experiment 1: High-Dimensional Weak-Support Mechanism Study}
\label{subsec:exp1}

\paragraph{Experimental setup.}
We construct a ten-dimensional regression problem with three Gaussian regions: dense low-noise, sparse low-noise, and dense high-noise. The two low-noise regions share the same observation-noise level, which separates the effect of training support from heteroscedasticity. The sparse low-noise region receives substantially less probability under the training distribution than under the calibration and test distributions. The conditional mean contains a localized nonlinear component around this region.

The experiment uses \(1{,}000\) training samples, \(2{,}000\) calibration samples, and \(10{,}000\) test samples. We compare Split CP, CV+, LACP, CQR, UACQR-S, EPICSCORE-MDN, LWCP, and CLAPS. For CV+, the original training set is retained in every fold model, while the calibration set is partitioned into five folds; each model uses four calibration folds for fitting and the remaining fold for residual evaluation. Consequently, each CV+ fold model uses additional labeled observations from the target-distribution calibration sample, whereas the split-conformal methods reserve this sample exclusively for calibration. The principal regional metrics are sparse-region coverage and the ratio of sparse-region width to dense low-noise width.

\paragraph{Results and analysis.}
Tables~\ref{tab:exp1_weak_support} and~\ref{tab:exp1_paired_lacp} show marginal coverage near the target for most methods, with slightly conservative coverage from CV+. CV+ attains the lowest interval score and the highest sparse-region coverage while maintaining a nearly uniform sparse-to-dense width ratio, reflecting its cross-fitted use of the target-distribution calibration sample. LACP, EPICSCORE-MDN, and CLAPS attain sparse-region coverage close to the target through stronger regional expansion, whereas Split CP, CQR, UACQR-S, and LWCP allocate less width to the sparse region and attain lower regional coverage. Relative to LACP, CLAPS maintains comparable marginal coverage and achieves a significantly lower interval score, together with slightly lower sparse-region coverage and a smaller sparse-to-dense width ratio.

\begin{table}[ht]
\centering
\caption{High-dimensional weak-support mechanism study. Sparse/Dense Width is the ratio of average interval width in the sparse low-noise region to that in the dense low-noise region.}
\label{tab:exp1_weak_support}
\small
\setlength{\tabcolsep}{5pt}
\begin{tabular}{lcccc}
\toprule
Method & Marginal Cov. & Sparse Cov. & Sparse/Dense Width & Interval Score \\
\midrule
Split CP & \(0.8986 \pm 0.0012\) & \(0.8402 \pm 0.0053\) & \(1.0000 \pm 0.0000\) & \(2.5451 \pm 0.0415\) \\
CV+ & \(0.9074 \pm 0.0012\) & \(0.9701 \pm 0.0014\) & \(1.0160 \pm 0.0010\) & \(1.9008 \pm 0.0068\) \\
LACP & \(0.8998 \pm 0.0012\) & \(0.9016 \pm 0.0101\) & \(4.4414 \pm 0.1890\) & \(2.6777 \pm 0.0536\) \\
CQR & \(0.8984 \pm 0.0014\) & \(0.8162 \pm 0.0053\) & \(1.8488 \pm 0.0520\) & \(2.8165 \pm 0.0348\) \\
UACQR-S & \(0.8981 \pm 0.0013\) & \(0.8246 \pm 0.0048\) & \(1.8924 \pm 0.0532\) & \(2.7876 \pm 0.0313\) \\
EPICSCORE-MDN & \(0.8995 \pm 0.0016\) & \(0.9057 \pm 0.0032\) & \(3.9642 \pm 0.1020\) & \(2.5340 \pm 0.0236\) \\
LWCP & \(0.9004 \pm 0.0012\) & \(0.7907 \pm 0.0055\) & \(1.0506 \pm 0.0025\) & \(3.1544 \pm 0.0456\) \\
CLAPS & \(0.8994 \pm 0.0011\) & \(0.8990 \pm 0.0103\) & \(4.3962 \pm 0.1863\) & \(2.6419 \pm 0.0537\) \\
\bottomrule
\end{tabular}
\end{table}

\begin{table}[ht]
\centering
\caption{Paired differences between CLAPS and LACP for the primary metrics.}
\label{tab:exp1_paired_lacp}
\small
\begin{tabular}{lcc}
\toprule
Metric & CLAPS $-$ LACP & 95\% CI \\
\midrule
Marginal Coverage & $-0.0004$ & $[-0.0012,\phantom{-}0.0004]$ \\
Sparse Coverage & $-0.0026$ & $[-0.0040,-0.0012]$ \\
Sparse/Dense Width Ratio & $-0.0451$ & $[-0.0561,-0.0345]$ \\
Interval Score & $-0.0357$ & $[-0.0424,-0.0292]$ \\
\bottomrule
\end{tabular}
\end{table}

\subsection{Experiment 2: Isolating the Noise-Weighted Feature Geometry}
\label{subsec:exp2}

\paragraph{Experimental setup.}
We use four Gaussian regions defined by the combination of dense or sparse training support and low or high observation noise. The calibration and test distributions are balanced across regions, whereas the training distribution concentrates most observations in the dense regions. The two sparse regions share the same localized nonlinear component in the conditional mean.

The experiment uses \(1{,}200\) training samples, \(2{,}500\) calibration samples, and \(12{,}000\) test samples. We compare LACP, Unweighted Laplace, LWCP, and CLAPS. Sparse-to-dense width ratios are computed separately within the low- and high-noise conditions, providing a direct comparison of support adaptation at a fixed observation-noise level.

\paragraph{Results and analysis.}
Table~\ref{tab:exp2_noise_weighted_geometry} shows comparable marginal coverage across all four methods. CLAPS produces a larger low-noise sparse-to-dense width ratio than Unweighted Laplace, while their high-noise ratios remain close; paired comparisons confirm both differences. CLAPS also improves the interval score relative to LACP, although Unweighted Laplace achieves the best aggregate score. LWCP maintains width ratios close to one and records the highest interval score.

\begin{table}[ht]
\centering
\caption{Isolation of noise-weighted feature geometry. The low- and high-noise columns report the ratio of average interval width in the sparse region to that in the dense region at the same observation-noise level.}
\label{tab:exp2_noise_weighted_geometry}
\small
\setlength{\tabcolsep}{4pt}
\begin{tabular}{lcccc}
\toprule
\multirow{2}{*}{Method}
& \multirow{2}{*}{Marginal Cov.}
& \multicolumn{2}{c}{Sparse/Dense Width}
& \multirow{2}{*}{Interval Score} \\
\cmidrule(lr){3-4}
& & Low-noise & High-noise & \\
\midrule
LACP
& \(0.9012 \pm 0.0011\)
& \(6.0382 \pm 0.2147\)
& \(1.9036 \pm 0.0559\)
& \(3.6657 \pm 0.1137\) \\
Unweighted Laplace
& \(0.9014 \pm 0.0012\)
& \(4.9728 \pm 0.1622\)
& \(1.8959 \pm 0.0547\)
& \(3.5967 \pm 0.1142\) \\
LWCP
& \(0.9012 \pm 0.0013\)
& \(1.0373 \pm 0.0017\)
& \(1.0352 \pm 0.0016\)
& \(3.9800 \pm 0.0846\) \\
CLAPS
& \(0.9010 \pm 0.0011\)
& \(5.9649 \pm 0.2094\)
& \(1.9090 \pm 0.0559\)
& \(3.6398 \pm 0.1121\) \\
\bottomrule
\end{tabular}
\end{table}

\subsection{Experiment 3: When Does the Epistemic Correction Matter?}
\label{subsec:exp3}

\paragraph{Experimental setup.}
We return to a three-region regression problem with dense low-noise, sparse low-noise, and dense high-noise regions. Training support varies only in the sparse low-noise region, and the calibration and test distributions remain fixed.

The first sweep varies the total training-set size over
\[
n_{\mathrm{tr}}
\in
\{100,250,500,1000,2000\}
\]
at a fixed sparse-region retention rate. The second sweep fixes the total training size and varies the sparse-region retention rate over
\[
r_{\mathrm{sparse}}
\in
\{1.00,0.20,0.10,0.05,0.025,0.01\}.
\]
The experiment uses \(2{,}000\) calibration samples and \(10{,}000\) test samples. We compare LACP, Unweighted Laplace, LWCP, and CLAPS using the sparse-to-dense width ratio, sparse-region interval score, and CLAPS epistemic fraction.

\begin{figure}[ht]
    \centering
    \includegraphics[width=\linewidth]{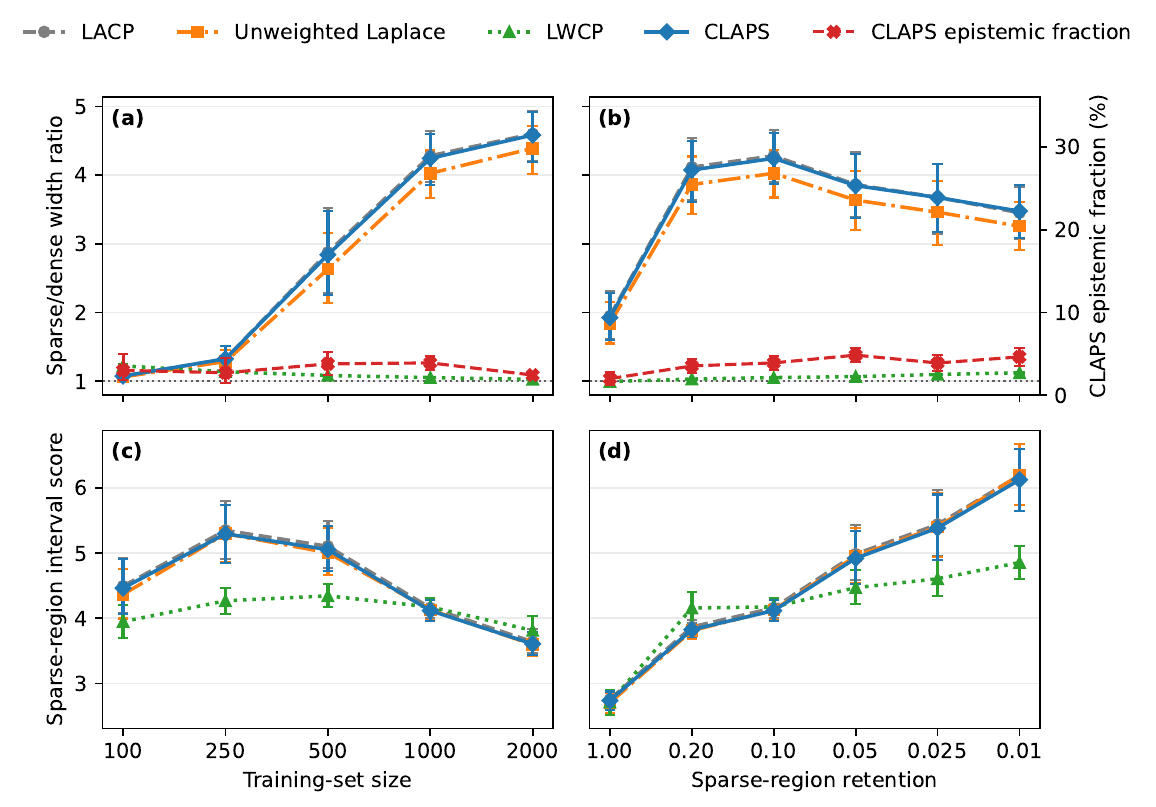}
    \caption{Regime study for the last-layer epistemic correction. Panels (a) and (c) vary the total training-set size at fixed sparse-region retention. Panels (b) and (d) vary sparse-region retention at fixed total training size. The top row reports the sparse-to-dense interval-width ratio and CLAPS epistemic fraction; the bottom row reports sparse-region interval score. Error bars show \(95\%\) bootstrap confidence intervals.}
    \label{fig:exp3_regime_study}
\end{figure}

\paragraph{Results and analysis.}
Figure~\ref{fig:exp3_regime_study} shows a nonmonotonic pattern across training sizes: the CLAPS epistemic fraction fluctuates at intermediate sizes and reaches its lowest value at the largest training size, while CLAPS and LACP retain similar regional width patterns and sparse-region scores. The local-support sweep shows a clearer response, with the epistemic fraction generally increasing as sparse-region retention decreases and the CLAPS--LACP score difference becoming larger under weaker support.

The sparse-to-dense width ratio increases over most of the support sweep but declines under the most severe thinning, where the sparse-region interval score also deteriorates. Unweighted Laplace remains competitive in sparse-region score, whereas LWCP exhibits little change in regional width.

\subsection{Experiment 4: Real-Data Benchmark and Support-Stratified Evaluation}
\label{subsec:exp4}

\paragraph{Experimental setup.}
We evaluate Concrete Compressive Strength, Energy Efficiency, Yacht Hydrodynamics, Airfoil Self-Noise, Wine Quality Red, Naval Propulsion, Bike Sharing, and California Housing. The benchmark compares Split CP, CV+, LACP, CQR, UACQR-S, EPICSCORE-MDN, LWCP, and CLAPS.

Response scales are standardized within each dataset. Width and interval-score effects are reported relative to LACP:
\[
\Delta_{\mathrm{width}}
=
\frac{\mathrm{Width}_{m}}
     {\mathrm{Width}_{\mathrm{LACP}}}-1,
\qquad
\Delta_{\mathrm{score}}
=
\frac{\mathrm{Score}_{m}}
     {\mathrm{Score}_{\mathrm{LACP}}}-1.
\]
Negative values indicate improved efficiency.

We measure support using the average distance to the \(20\) nearest training observations in both the standardized raw-input space and the learned feature space. Test points are divided into five equally sized groups, with Q1 denoting the strongest support and Q5 the weakest support. Table~\ref{tab:exp4_real_data} reports the aggregate benchmark results and the interval-score effect in the weakest learned-feature support quintile. Tables~\ref{tab:exp4_hierarchical_paired} and \ref{tab:exp4_feature_quintiles} provide the hierarchical paired comparison with LACP and the complete learned-feature support-stratified results, respectively. The corresponding raw-input analysis is reported in the appendix.

\paragraph{Results and analysis.}
Table~\ref{tab:exp4_real_data} shows coverage close to the target level across methods. CLAPS achieves modest reductions in average width and interval score relative to LACP, while EPICSCORE-MDN is the closest competing method under the evaluated implementations. The hierarchical paired results in Table~\ref{tab:exp4_hierarchical_paired} show confidence intervals below zero for both width and interval score, whereas the coverage interval includes zero.

Table~\ref{tab:exp4_feature_quintiles} shows narrower CLAPS intervals across Q1--Q4. In the weakest-support quintile, Q5, the average width remains nearly unchanged, coverage increases, and the interval-score improvement is largest. The raw-input support analysis in the appendix exhibits a similar overall pattern.

\begin{table}[ht]
\centering
\caption{Real-data benchmark over eight datasets. Width and interval-score changes are relative to LACP. Weak-Q5 Score is the relative interval-score change in the weakest learned-feature support quintile.}
\label{tab:exp4_real_data}
\small
\setlength{\tabcolsep}{3.5pt}
\begin{tabular}{lcccc}
\toprule
\multirow{2}{*}{Method}
& \multirow{2}{*}{Coverage}
& \multicolumn{2}{c}{Comparison with LACP}
& \multirow{2}{*}{Weak-Q5 Score} \\
\cmidrule(lr){3-4}
& & \(\Delta\) Width & \(\Delta\) Score & \\
\midrule
Split CP
& \(0.9017 \pm 0.0021\)
& \(16.80 \pm 2.91\%\)
& \(25.53 \pm 3.04\%\)
& \(18.34 \pm 3.04\%\) \\
CV+
& \(0.9164 \pm 0.0021\)
& \(18.02 \pm 2.76\%\)
& \(18.21 \pm 2.29\%\)
& \(\phantom{-}8.79 \pm 2.47\%\) \\
LACP
& \(0.8998 \pm 0.0021\)
& \(\phantom{-}0.00 \pm 0.00\%\)
& \(\phantom{-}0.00 \pm 0.00\%\)
& \(\phantom{-}0.00 \pm 0.00\%\) \\
CQR
& \(0.9000 \pm 0.0022\)
& \(40.24 \pm 5.84\%\)
& \(31.94 \pm 4.79\%\)
& \(30.43 \pm 4.26\%\) \\
UACQR-S
& \(0.8989 \pm 0.0023\)
& \(35.54 \pm 4.97\%\)
& \(37.27 \pm 5.92\%\)
& \(39.65 \pm 6.31\%\) \\
EPICSCORE-MDN
& \(0.9007 \pm 0.0026\)
& \(\phantom{-}7.61 \pm 4.00\%\)
& \(\phantom{-}4.95 \pm 2.84\%\)
& \(14.66 \pm 10.59\%\) \\
LWCP
& \(0.9004 \pm 0.0023\)
& \(18.20 \pm 2.68\%\)
& \(28.51 \pm 2.36\%\)
& \(23.82 \pm 2.65\%\) \\
CLAPS
& \(0.8983 \pm 0.0020\)
& \(-1.30 \pm 0.26\%\)
& \(-1.06 \pm 0.19\%\)
& \(-1.75 \pm 0.28\%\) \\
\bottomrule
\end{tabular}
\end{table}

\begin{table}[ht]
\centering
\caption{Paired comparisons between CLAPS and LACP on the real-data benchmarks. All differences are computed as CLAPS minus LACP.}
\label{tab:exp4_paired_support}

\begin{subtable}[t]{\linewidth}
\centering
\caption{Hierarchical paired bootstrap comparison.}
\label{tab:exp4_hierarchical_paired}
\small
\setlength{\tabcolsep}{6pt}
\begin{tabular}{lcc}
\toprule
Effect & Mean difference & 95\% CI \\
\midrule
Coverage
& $-0.0015$
& $[-0.0046,\phantom{-}0.0005]$ \\

Width
& $-1.3000\%$
& $[-2.8712\%,-0.3143\%]$ \\

Interval score
& $-1.0579\%$
& $[-2.3197\%,-0.3283\%]$ \\

Weak-Q5 interval score
& $-1.7531\%$
& $[-3.0600\%,-0.7357\%]$ \\
\bottomrule
\end{tabular}
\end{subtable}

\vspace{0.7em}

\begin{subtable}[t]{\linewidth}
\centering
\caption{Differences across learned-feature support quintiles. Q1 and Q5 denote the strongest and weakest support, respectively.}
\label{tab:exp4_feature_quintiles}
\small
\setlength{\tabcolsep}{6pt}
\begin{tabular}{lccc}
\toprule
Quintile
& \(\Delta\) Coverage
& \(\Delta\) Width
& \(\Delta\) Score \\
\midrule
Q1
& $-0.0021 \pm 0.0006$
& $-1.66 \pm 0.24\%$
& $-0.84 \pm 0.21\%$ \\

Q2
& $-0.0037 \pm 0.0012$
& $-1.26 \pm 0.23\%$
& $-0.62 \pm 0.21\%$ \\

Q3
& $-0.0026 \pm 0.0018$
& $-1.20 \pm 0.21\%$
& $-0.36 \pm 0.15\%$ \\

Q4
& $-0.0042 \pm 0.0018$
& $-1.29 \pm 0.24\%$
& $-0.70 \pm 0.22\%$ \\

Q5
& $\phantom{-}0.0052 \pm 0.0028$
& $\phantom{-}0.05 \pm 0.34\%$
& $-1.75 \pm 0.28\%$ \\
\bottomrule
\end{tabular}
\end{subtable}
\end{table}

\section{Discussion}
\label{sec:discussion}

\subsection{Noise-Weighted Geometry and Conformal Scaling}
\label{subsec:noise_weighted_geometry}

CLAPS uses standard split conformal calibration to obtain finite-sample marginal coverage after the score function has been fixed. Its methodological contribution lies in the predictive scale: the learned heteroscedastic variance determines the local observation scale and weights each training feature's contribution to the last-layer posterior precision. This dual role produces a noise-weighted feature geometry that reflects both response variability and reliable training support.

The conformal quantile sets the overall calibration level, and the predictive scale distributes interval width across inputs. High estimated observation variance enlarges the aleatoric component, whereas limited support from low-noise training features increases the last-layer correction. CLAPS thereby provides a structured local scaling rule within split conformal regression.

\subsection{Relation to Aleatoric and Epistemic Baselines}
\label{subsec:relation_baselines}

LACP is the closest matched baseline because it uses the same heteroscedastic regressor without the last-layer correction. CLAPS is therefore best interpreted as a targeted extension of aleatoric local scaling. The methodological contribution centers on the heteroscedastic neural-linear scale, its noise-weighted posterior geometry, and its regime-dependent interval allocation under standard conformal calibration. Unweighted Laplace isolates the effect of heteroscedastic precision weighting, LWCP represents leverage-based adaptation, and UACQR-S and EPICSCORE-MDN provide broader epistemic-aware comparisons.

The experiments reveal method- and regime-dependent trade-offs. CLAPS generally preserves the allocation pattern of LACP and provides modest efficiency gains, whereas EPICSCORE-MDN and Unweighted Laplace remain competitive in several aggregate comparisons. The main empirical distinction of CLAPS is the behavior of its noise-weighted correction under controlled support imbalance and its paired improvements over LACP on real data.

\subsection{When the Epistemic Correction Matters}
\label{subsec:epistemic_correction}

The last-layer correction becomes most influential when training support is uneven across the learned representation. The support-retention experiments show a clearer increase in the correction than the global training-size sweep, indicating that localized scarcity is more important than sample size alone. This effect is strongest when weak support occurs in regions with relatively informative, low-noise observations.

As weighted feature information increases, the posterior covariance contracts and CLAPS approaches aleatoric local scaling. Extreme data scarcity can also weaken the learned representation and variance head, reducing the reliability of the resulting allocation. CLAPS is therefore most relevant when localized support imbalance coexists with a sufficiently informative learned representation.

\subsection{Practical and Computational Implications}
\label{subsec:practical_implications}

The two variance components provide a model-based diagnostic of how uncertainty is allocated. The observation component identifies regions assigned a large heteroscedastic scale, and the last-layer component identifies representations receiving a large support correction. These quantities can guide data collection, human review, and cautious downstream decisions, although the learned variance head may also reflect model error or representation quality.

The prior precision controls the strength of the last-layer correction: larger values shrink CLAPS toward LACP, and smaller values increase the influence of posterior uncertainty. Training-only inner validation selects this value with the fitted regressor held fixed and keeps calibration and test responses outside the selection procedure. The covariance is constructed once after fitting, and each test input adds a last-layer quadratic-form evaluation after feature extraction. Appendix~\ref{app:feature_dimension} reports this additional cost.

\subsection{Limitations and Future Work}
\label{subsec:limitations}

CLAPS inherits the standard scope of split conformal prediction. Its guarantee is marginal and assumes exchangeability between calibration and test examples, so regional and support-stratified coverage remain empirical properties. Extensions to covariate shift, temporal dependence, and stronger subgroup-level guarantees are important directions for future work.

The epistemic component is conditional on the fixed learned representation and does not capture uncertainty over the full neural network. A severely shifted input may be mapped to a familiar latent region and consequently receive a small last-layer correction. Future work could incorporate representation-aware uncertainty, richer posterior approximations, more robust variance modeling, and scalable covariance approximations for high-dimensional final layers.

\section{Conclusion}
\label{sec:conclusion}

We presented CLAPS, a split conformal regression method that combines a learned heteroscedastic observation scale with a last-layer Laplace support correction. Its main structural feature is a noise-weighted feature geometry in which the learned variance controls both test-time scaling and the contribution of training features to posterior precision. Theoretical analysis characterizes this construction and its reduction toward aleatoric local scaling as weighted feature information increases. Controlled experiments show that the correction is most relevant under localized support imbalance, and real-data results demonstrate modest paired efficiency improvements over the matched LACP baseline with comparable marginal coverage. These findings support heteroscedastic last-layer uncertainty as a practical mechanism for support-aware interval allocation within a learned representation.

\section*{Broader Impact Statement}

This work studies prediction intervals for regression and is primarily methodological. Its potential positive impact is to make conformal prediction intervals more informative in heterogeneous settings by distinguishing interval width due to estimated observation noise from width due to limited training support. Such information may help practitioners identify regions where uncertainty is driven by irreducible variability and regions where additional data collection, human review, or more cautious decisions may be appropriate.

At the same time, the guarantees considered in this work are marginal and rely on exchangeability between calibration and test examples. They should not be interpreted as conditional coverage guarantees for every input, subgroup, or deployment environment. In high-stakes domains such as healthcare, finance, public policy, or employment, prediction intervals produced by CLAPS should not be used as the sole basis for automated decisions. Practical deployment would require domain-specific validation, subgroup and distribution-shift audits, and careful consideration of how uncertainty estimates affect downstream users and affected individuals.

\section*{Reproducibility}

To support reproducibility, we provide a supplementary notebook, \texttt{CLAPS.ipynb}, containing the implementation of CLAPS and all experiments reported in the paper. The notebook includes the synthetic studies, real-data benchmark, baseline comparisons, evaluation metrics, and table-generation code used to produce the experimental results.

\bibliography{main}

@inproceedings{saunders1999transduction,
  title={Transduction with confidence and credibility},
  author={Saunders, C and Gammerman, A and Vovk, V},
  booktitle={Proceedings of the 16th international joint conference on Artificial intelligence-Volume 2},
  pages={722--726},
  year={1999}
}

@inproceedings{papadopoulos2002inductive,
  title={Inductive confidence machines for regression},
  author={Papadopoulos, Harris and Proedrou, Kostas and Vovk, Volodya and Gammerman, Alex},
  booktitle={European conference on machine learning},
  pages={345--356},
  year={2002},
  organization={Springer}
}

@article{gammerman2007hedging,
  title={Hedging predictions in machine learning},
  author={Gammerman, Alexander and Vovk, Vladimir},
  journal={The Computer Journal},
  volume={50},
  number={2},
  pages={151--163},
  year={2007},
  publisher={OUP}
}

@article{shafer2008tutorial,
  title={A tutorial on conformal prediction.},
  author={Shafer, Glenn and Vovk, Vladimir},
  journal={Journal of machine learning research},
  volume={9},
  number={3},
  year={2008}
}

@article{papadopoulos2011regression,
  title={Regression conformal prediction with nearest neighbours},
  author={Papadopoulos, Harris and Vovk, Vladimir and Gammerman, Alex},
  journal={Journal of Artificial Intelligence Research},
  volume={40},
  pages={815--840},
  year={2011}
}

@article{johansson2014regression,
  title={Regression conformal prediction with random forests},
  author={Johansson, Ulf and Bostr{\"o}m, Henrik and L{\"o}fstr{\"o}m, Tuve and Linusson, Henrik},
  journal={Machine learning},
  volume={97},
  number={1},
  pages={155--176},
  year={2014},
  publisher={Springer}
}

@article{lei2018distribution,
  title={Distribution-free predictive inference for regression},
  author={Lei, Jing and G’Sell, Max and Rinaldo, Alessandro and Tibshirani, Ryan J and Wasserman, Larry},
  journal={Journal of the American Statistical Association},
  volume={113},
  number={523},
  pages={1094--1111},
  year={2018},
  publisher={Taylor \& Francis}
}

@article{ndiaye2019computing,
  title={Computing full conformal prediction set with approximate homotopy},
  author={Ndiaye, Eugene and Takeuchi, Ichiro},
  journal={Advances in Neural Information Processing Systems},
  volume={32},
  year={2019}
}

@article{barber2021predictive,
  title={Predictive inference with the jackknife+},
  author={Barber, Rina Foygel and Candes, Emmanuel J and Ramdas, Aaditya and Tibshirani, Ryan J},
  journal={The Annals of Statistics},
  volume={49},
  number={1},
  pages={486--507},
  year={2021},
  publisher={JSTOR}
}

@article{steinberger2023conditional,
  title={Conditional predictive inference for stable algorithms},
  author={Steinberger, Lukas and Leeb, Hannes},
  journal={The Annals of Statistics},
  volume={51},
  number={1},
  pages={290--311},
  year={2023},
  publisher={Institute of Mathematical Statistics}
}

@article{romano2019conformalized,
  title={Conformalized quantile regression},
  author={Romano, Yaniv and Patterson, Evan and Candes, Emmanuel},
  journal={Advances in neural information processing systems},
  volume={32},
  year={2019}
}

@article{tibshirani2019conformal,
  title={Conformal prediction under covariate shift},
  author={Tibshirani, Ryan J and Foygel Barber, Rina and Candes, Emmanuel and Ramdas, Aaditya},
  journal={Advances in neural information processing systems},
  volume={32},
  year={2019}
}

@article{foygel2021limits,
  title={The limits of distribution-free conditional predictive inference},
  author={Foygel Barber, Rina and Candes, Emmanuel J and Ramdas, Aaditya and Tibshirani, Ryan J},
  journal={Information and Inference: A Journal of the IMA},
  volume={10},
  number={2},
  pages={455--482},
  year={2021},
  publisher={Oxford University Press}
}

@article{gibbs2021adaptive,
  title={Adaptive conformal inference under distribution shift},
  author={Gibbs, Isaac and Candes, Emmanuel},
  journal={Advances in Neural Information Processing Systems},
  volume={34},
  pages={1660--1672},
  year={2021}
}

@article{barber2023conformal,
  title={Conformal prediction beyond exchangeability},
  author={Barber, Rina Foygel and Candes, Emmanuel J and Ramdas, Aaditya and Tibshirani, Ryan J},
  journal={The Annals of Statistics},
  volume={51},
  number={2},
  pages={816--845},
  year={2023},
  publisher={Institute of Mathematical Statistics}
}

@article{bian2023training,
  title={Training-conditional coverage for distribution-free predictive inference},
  author={Bian, Michael and Barber, Rina Foygel},
  journal={Electronic Journal of Statistics},
  volume={17},
  number={2},
  pages={2044--2066},
  year={2023},
  publisher={The Institute of Mathematical Statistics and the Bernoulli Society}
}

@article{gibbs2024conformal,
  title={Conformal inference for online prediction with arbitrary distribution shifts},
  author={Gibbs, Isaac and Cand{\`e}s, Emmanuel J},
  journal={Journal of Machine Learning Research},
  volume={25},
  number={162},
  pages={1--36},
  year={2024}
}

@article{oliveira2024split,
  title={Split conformal prediction and non-exchangeable data},
  author={Oliveira, Roberto I and Orenstein, Paulo and Ramos, Thiago and Romano, Joao Vitor},
  journal={Journal of Machine Learning Research},
  volume={25},
  number={225},
  pages={1--38},
  year={2024}
}

@inproceedings{papadopoulos2008normalized,
  title={Normalized nonconformity measures for regression conformal prediction},
  author={Papadopoulos, Harris and Gammerman, Alex and Vovk, Volodya},
  booktitle={Proceedings of the IASTED International Conference on Artificial Intelligence and Applications (AIA 2008)},
  pages={64--69},
  year={2008}
}

@inproceedings{seedat2023improving,
  title={Improving adaptive conformal prediction using self-supervised learning},
  author={Seedat, Nabeel and Jeffares, Alan and Imrie, Fergus and van der Schaar, Mihaela},
  booktitle={International Conference on Artificial Intelligence and Statistics},
  pages={10160--10177},
  year={2023},
  organization={PMLR}
}

@article{guan2023localized,
  title={Localized conformal prediction: A generalized inference framework for conformal prediction},
  author={Guan, Leying},
  journal={Biometrika},
  volume={110},
  number={1},
  pages={33--50},
  year={2023},
  publisher={Oxford University Press}
}

@article{hore2025conformal,
  title={Conformal prediction with local weights: randomization enables robust guarantees},
  author={Hore, Rohan and Barber, Rina Foygel},
  journal={Journal of the Royal Statistical Society Series B: Statistical Methodology},
  volume={87},
  number={2},
  pages={549--578},
  year={2025},
  publisher={Oxford University Press UK}
}

@article{gibbs2025conformal,
  title={Conformal prediction with conditional guarantees},
  author={Gibbs, Isaac and Cherian, John J and Cand{\`e}s, Emmanuel J},
  journal={Journal of the Royal Statistical Society Series B: Statistical Methodology},
  volume={87},
  number={4},
  pages={1100--1126},
  year={2025},
  publisher={Oxford University Press UK}
}

@article{jung2022batch,
  title={Batch multivalid conformal prediction},
  author={Jung, Christopher and Noarov, Georgy and Ramalingam, Ramya and Roth, Aaron},
  journal={arXiv preprint arXiv:2209.15145},
  year={2022}
}

@article{bastani2022practical,
  title={Practical adversarial multivalid conformal prediction},
  author={Bastani, Osbert and Gupta, Varun and Jung, Christopher and Noarov, Georgy and Ramalingam, Ramya and Roth, Aaron},
  journal={Advances in neural information processing systems},
  volume={35},
  pages={29362--29373},
  year={2022}
}

@article{kiyani2024conformal,
  title={Conformal prediction with learned features},
  author={Kiyani, Shayan and Pappas, George and Hassani, Hamed},
  journal={arXiv preprint arXiv:2404.17487},
  year={2024}
}

@article{kiyani2024length,
  title={Length optimization in conformal prediction},
  author={Kiyani, Shayan and Pappas, George and Hassani, Hamed},
  journal={Advances in Neural Information Processing Systems},
  volume={37},
  pages={99519--99563},
  year={2024}
}

@article{chernozhukov2021distributional,
  title={Distributional conformal prediction},
  author={Chernozhukov, Victor and W{\"u}thrich, Kaspar and Zhu, Yinchu},
  journal={Proceedings of the National Academy of Sciences},
  volume={118},
  number={48},
  pages={e2107794118},
  year={2021},
  publisher={National Academy of Sciences}
}

@article{sesia2021conformal,
  title={Conformal prediction using conditional histograms},
  author={Sesia, Matteo and Romano, Yaniv},
  journal={Advances in neural information processing systems},
  volume={34},
  pages={6304--6315},
  year={2021}
}

@inproceedings{wang2023probabilistic,
  title={Probabilistic Conformal Prediction Using Conditional Random Samples},
  author={Wang, Zhendong and Gao, Ruijiang and Yin, Mingzhang and Zhou, Mingyuan and Blei, David},
  booktitle={International Conference on Artificial Intelligence and Statistics},
  pages={8814--8836},
  year={2023},
  organization={PMLR}
}

@article{plassier2024probabilistic,
  title={Probabilistic conformal prediction with approximate conditional validity},
  author={Plassier, Vincent and Fishkov, Alexander and Guizani, Mohsen and Panov, Maxim and Moulines, Eric},
  journal={arXiv preprint arXiv:2407.01794},
  year={2024}
}

@article{guha2024conformal,
  title={Conformal prediction via regression-as-classification},
  author={Guha, Etash and Natarajan, Shlok and M{\"o}llenhoff, Thomas and Khan, Mohammad Emtiyaz and Ndiaye, Eugene},
  journal={arXiv preprint arXiv:2404.08168},
  year={2024}
}

@inproceedings{colombo2024normalizing,
  title={Normalizing flows for conformai regression},
  author={Colombo, Nicolo},
  booktitle={Proceedings of the Fortieth Conference on Uncertainty in Artificial Intelligence},
  pages={881--893},
  year={2024}
}

@inproceedings{tumu2024multi,
  title={Multi-modal conformal prediction regions by optimizing convex shape templates},
  author={Tumu, Renukanandan and Cleaveland, Matthew and Mangharam, Rahul and Pappas, George and Lindemann, Lars},
  booktitle={6th Annual Learning for Dynamics \& Control Conference},
  pages={1343--1356},
  year={2024},
  organization={PMLR}
}

@inproceedings{plassier2025rectifying,
  title={Rectifying conformity scores for better conditional coverage},
  author={Plassier, Vincent and Fishkov, Alexander and Dheur, Victor and Guizani, Mohsen and Ben Taieb, Souhaib and Panov, Maxim and Moulines, Eric},
  booktitle={The 42nd International Conference on Machine Learning},
  year={2025},
  organization={PMLR}
}

@article{van2024self,
  title={Self-calibrating conformal prediction},
  author={van der Laan, Lars and Alaa, Ahmed M},
  journal={Advances in Neural Information Processing Systems},
  volume={37},
  pages={107138--107170},
  year={2024}
}

@article{romano2020classification,
  title={Classification with valid and adaptive coverage},
  author={Romano, Yaniv and Sesia, Matteo and Candes, Emmanuel},
  journal={Advances in neural information processing systems},
  volume={33},
  pages={3581--3591},
  year={2020}
}

@article{angelopoulos2020uncertainty,
  title={Uncertainty sets for image classifiers using conformal prediction},
  author={Angelopoulos, Anastasios and Bates, Stephen and Malik, Jitendra and Jordan, Michael I},
  journal={arXiv preprint arXiv:2009.14193},
  year={2020}
}

@article{stutz2021learning,
  title={Learning optimal conformal classifiers},
  author={Stutz, David and Cemgil, Ali Taylan and Doucet, Arnaud and others},
  journal={arXiv preprint arXiv:2110.09192},
  year={2021}
}

@inproceedings{huang2024conformal,
  title={Conformal Prediction for Deep Classifier via Label Ranking},
  author={Huang, Jianguo and Xi, Huajun and Zhang, Linjun and Yao, Huaxiu and Qiu, Yue and Wei, Hongxin},
  booktitle={International Conference on Machine Learning},
  pages={20331--20347},
  year={2024},
  organization={PMLR}
}

@article{graves2011practical,
  title={Practical variational inference for neural networks},
  author={Graves, Alex},
  journal={Advances in neural information processing systems},
  volume={24},
  year={2011}
}

@inproceedings{hernandez2015probabilistic,
  title={Probabilistic backpropagation for scalable learning of bayesian neural networks},
  author={Hern{\'a}ndez-Lobato, Jos{\'e} Miguel and Adams, Ryan},
  booktitle={International conference on machine learning},
  pages={1861--1869},
  year={2015},
  organization={PMLR}
}

@inproceedings{blundell2015weight,
  title={Weight uncertainty in neural network},
  author={Blundell, Charles and Cornebise, Julien and Kavukcuoglu, Koray and Wierstra, Daan},
  booktitle={International conference on machine learning},
  pages={1613--1622},
  year={2015},
  organization={PMLR}
}

@article{kingma2015variational,
  title={Variational dropout and the local reparameterization trick},
  author={Kingma, Durk P and Salimans, Tim and Welling, Max},
  journal={Advances in neural information processing systems},
  volume={28},
  year={2015}
}

@inproceedings{gal2016dropout,
  title={Dropout as a bayesian approximation: Representing model uncertainty in deep learning},
  author={Gal, Yarin and Ghahramani, Zoubin},
  booktitle={international conference on machine learning},
  pages={1050--1059},
  year={2016},
  organization={PMLR}
}

@inproceedings{louizos2017multiplicative,
  title={Multiplicative normalizing flows for variational bayesian neural networks},
  author={Louizos, Christos and Welling, Max},
  booktitle={International conference on machine learning},
  pages={2218--2227},
  year={2017},
  organization={PMLR}
}

@article{gal2017concrete,
  title={Concrete dropout},
  author={Gal, Yarin and Hron, Jiri and Kendall, Alex},
  journal={Advances in neural information processing systems},
  volume={30},
  year={2017}
}

@article{krueger2017bayesian,
  title={Bayesian hypernetworks},
  author={Krueger, David and Huang, Chin-Wei and Islam, Riashat and Turner, Ryan and Lacoste, Alexandre and Courville, Aaron},
  journal={arXiv preprint arXiv:1710.04759},
  year={2017}
}

@article{sun2019functional,
  title={Functional variational Bayesian neural networks},
  author={Sun, Shengyang and Zhang, Guodong and Shi, Jiaxin and Grosse, Roger},
  journal={arXiv preprint arXiv:1903.05779},
  year={2019}
}

@article{tran2019bayesian,
  title={Bayesian layers: A module for neural network uncertainty},
  author={Tran, Dustin and Dusenberry, Mike and Van Der Wilk, Mark and Hafner, Danijar},
  journal={Advances in neural information processing systems},
  volume={32},
  year={2019}
}

@article{lakshminarayanan2017simple,
  title={Simple and scalable predictive uncertainty estimation using deep ensembles},
  author={Lakshminarayanan, Balaji and Pritzel, Alexander and Blundell, Charles},
  journal={Advances in neural information processing systems},
  volume={30},
  year={2017}
}

@article{sensoy2018evidential,
  title={Evidential deep learning to quantify classification uncertainty},
  author={Sensoy, Murat and Kaplan, Lance and Kandemir, Melih},
  journal={Advances in neural information processing systems},
  volume={31},
  year={2018}
}

@article{amini2020deep,
  title={Deep evidential regression},
  author={Amini, Alexander and Schwarting, Wilko and Soleimany, Ava and Rus, Daniela},
  journal={Advances in neural information processing systems},
  volume={33},
  pages={14927--14937},
  year={2020}
}

@inproceedings{van2020uncertainty,
  title={Uncertainty estimation using a single deep deterministic neural network},
  author={Van Amersfoort, Joost and Smith, Lewis and Teh, Yee Whye and Gal, Yarin},
  booktitle={International conference on machine learning},
  pages={9690--9700},
  year={2020},
  organization={PMLR}
}

@article{charpentier2020posterior,
  title={Posterior network: Uncertainty estimation without ood samples via density-based pseudo-counts},
  author={Charpentier, Bertrand and Z{\"u}gner, Daniel and G{\"u}nnemann, Stephan},
  journal={Advances in neural information processing systems},
  volume={33},
  pages={1356--1367},
  year={2020}
}

@article{kendall2017uncertainties,
  title={What uncertainties do we need in bayesian deep learning for computer vision?},
  author={Kendall, Alex and Gal, Yarin},
  journal={Advances in neural information processing systems},
  volume={30},
  year={2017}
}

@inproceedings{kuleshov2018accurate,
  title={Accurate uncertainties for deep learning using calibrated regression},
  author={Kuleshov, Volodymyr and Fenner, Nathan and Ermon, Stefano},
  booktitle={International conference on machine learning},
  pages={2796--2804},
  year={2018},
  organization={PMLR}
}

@article{ovadia2019can,
  title={Can you trust your model's uncertainty? evaluating predictive uncertainty under dataset shift},
  author={Ovadia, Yaniv and Fertig, Emily and Ren, Jie and Nado, Zachary and Sculley, David and Nowozin, Sebastian and Dillon, Joshua and Lakshminarayanan, Balaji and Snoek, Jasper},
  journal={Advances in neural information processing systems},
  volume={32},
  year={2019}
}

@article{maddox2019simple,
  title={A simple baseline for bayesian uncertainty in deep learning},
  author={Maddox, Wesley J and Izmailov, Pavel and Garipov, Timur and Vetrov, Dmitry P and Wilson, Andrew Gordon},
  journal={Advances in neural information processing systems},
  volume={32},
  year={2019}
}

@article{wilson2020bayesian,
  title={Bayesian deep learning and a probabilistic perspective of generalization},
  author={Wilson, Andrew G and Izmailov, Pavel},
  journal={Advances in neural information processing systems},
  volume={33},
  pages={4697--4708},
  year={2020}
}

@inproceedings{ritter2018scalable,
  title={A scalable laplace approximation for neural networks},
  author={Ritter, Hippolyt and Botev, Aleksandar and Barber, David},
  booktitle={International conference on learning representations},
  year={2018}
}

@article{daxberger2021laplace,
  title={Laplace redux-effortless bayesian deep learning},
  author={Daxberger, Erik and Kristiadi, Agustinus and Immer, Alexander and Eschenhagen, Runa and Bauer, Matthias and Hennig, Philipp},
  journal={Advances in neural information processing systems},
  volume={34},
  pages={20089--20103},
  year={2021}
}

@inproceedings{kristiadi2021learnable,
  title={Learnable uncertainty under laplace approximations},
  author={Kristiadi, Agustinus and Hein, Matthias and Hennig, Philipp},
  booktitle={Uncertainty in Artificial Intelligence},
  pages={344--353},
  year={2021},
  organization={PMLR}
}

@inproceedings{immer2021scalable,
  title={Scalable marginal likelihood estimation for model selection in deep learning},
  author={Immer, Alexander and Bauer, Matthias and Fortuin, Vincent and R{\"a}tsch, Gunnar and Emtiyaz, Khan Mohammad},
  booktitle={International Conference on Machine Learning},
  pages={4563--4573},
  year={2021},
  organization={PMLR}
}

@article{cinquin2024fsp,
  title={FSP-Laplace: Function-space priors for the Laplace approximation in Bayesian deep learning},
  author={Cinquin, Tristan and Pf{\"o}rtner, Marvin and Fortuin, Vincent and Hennig, Philipp and Bamler, Robert},
  journal={Advances in Neural Information Processing Systems},
  volume={37},
  pages={13897--13926},
  year={2024}
}

@inproceedings{immer2021improving,
  title={Improving predictions of Bayesian neural nets via local linearization},
  author={Immer, Alexander and Korzepa, Maciej and Bauer, Matthias},
  booktitle={International conference on artificial intelligence and statistics},
  pages={703--711},
  year={2021},
  organization={PMLR}
}

@inproceedings{antoran2022adapting,
  title={Adapting the linearised laplace model evidence for modern deep learning},
  author={Antor{\'a}n, Javier and Janz, David and Allingham, James U and Daxberger, Erik and Barbano, Riccardo Rb and Nalisnick, Eric and Hern{\'a}ndez-Lobato, Jos{\'e} Miguel},
  booktitle={International Conference on Machine Learning},
  pages={796--821},
  year={2022},
  organization={PMLR}
}

@inproceedings{daxberger2021bayesian,
  title={Bayesian deep learning via subnetwork inference},
  author={Daxberger, Erik and Nalisnick, Eric and Allingham, James U and Antor{\'a}n, Javier and Hern{\'a}ndez-Lobato, Jos{\'e} Miguel},
  booktitle={International Conference on Machine Learning},
  pages={2510--2521},
  year={2021},
  organization={PMLR}
}

@article{deng2022accelerated,
  title={Accelerated linearized Laplace approximation for Bayesian deep learning},
  author={Deng, Zhijie and Zhou, Feng and Zhu, Jun},
  journal={Advances in Neural Information Processing Systems},
  volume={35},
  pages={2695--2708},
  year={2022}
}

@article{ortega2023variational,
  title={Variational linearized Laplace approximation for Bayesian deep learning},
  author={Ortega, Luis A and Santana, Sim{\'o}n Rodr{\'\i}guez and Hern{\'a}ndez-Lobato, Daniel},
  journal={arXiv preprint arXiv:2302.12565},
  year={2023}
}

@article{khan2019approximate,
  title={Approximate inference turns deep networks into Gaussian processes},
  author={Khan, Mohammad Emtiyaz and Immer, Alexander and Abedi, Ehsan and Korzepa, Maciej},
  journal={Advances in neural information processing systems},
  volume={32},
  year={2019}
}

@article{jacot2018neural,
  title={Neural tangent kernel: Convergence and generalization in neural networks},
  author={Jacot, Arthur and Gabriel, Franck and Hongler, Cl{\'e}ment},
  journal={Advances in neural information processing systems},
  volume={31},
  year={2018}
}

@inproceedings{maddox2021fast,
  title={Fast adaptation with linearized neural networks},
  author={Maddox, Wesley and Tang, Shuai and Moreno, Pablo and Wilson, Andrew Gordon and Damianou, Andreas},
  booktitle={International Conference on Artificial Intelligence and Statistics},
  pages={2737--2745},
  year={2021},
  organization={PMLR}
}

@inproceedings{kristiadi2020being,
  title={Being bayesian, even just a bit, fixes overconfidence in relu networks},
  author={Kristiadi, Agustinus and Hein, Matthias and Hennig, Philipp},
  booktitle={International conference on machine learning},
  pages={5436--5446},
  year={2020},
  organization={PMLR}
}

@inproceedings{watson2021latent,
  title={Latent derivative Bayesian last layer networks},
  author={Watson, Joe and Lin, Jihao Andreas and Klink, Pascal and Pajarinen, Joni and Peters, Jan},
  booktitle={International Conference on Artificial Intelligence and Statistics},
  pages={1198--1206},
  year={2021},
  organization={PMLR}
}

@article{harrison2024variational,
  title={Variational Bayesian last layers},
  author={Harrison, James and Willes, John and Snoek, Jasper},
  journal={arXiv preprint arXiv:2404.11599},
  year={2024}
}

@article{brunzema2024bayesian,
  title={Bayesian optimization via continual variational last layer training},
  author={Brunzema, Paul and Jordahn, Mikkel and Willes, John and Trimpe, Sebastian and Snoek, Jasper and Harrison, James},
  journal={arXiv preprint arXiv:2412.09477},
  year={2024}
}

@inproceedings{martens2015optimizing,
  title={Optimizing neural networks with kronecker-factored approximate curvature},
  author={Martens, James and Grosse, Roger},
  booktitle={International conference on machine learning},
  pages={2408--2417},
  year={2015},
  organization={PMLR}
}

@inproceedings{botev2017practical,
  title={Practical Gauss-Newton optimisation for deep learning},
  author={Botev, Aleksandar and Ritter, Hippolyt and Barber, David},
  booktitle={International Conference on Machine Learning},
  pages={557--565},
  year={2017},
  organization={PMLR}
}

@article{dangel2019backpack,
  title={Backpack: Packing more into backprop},
  author={Dangel, Felix and Kunstner, Frederik and Hennig, Philipp},
  journal={arXiv preprint arXiv:1912.10985},
  year={2019}
}

@article{eschenhagen2023kronecker,
  title={Kronecker-factored approximate curvature for modern neural network architectures},
  author={Eschenhagen, Runa and Immer, Alexander and Turner, Richard and Schneider, Frank and Hennig, Philipp},
  journal={Advances in Neural Information Processing Systems},
  volume={36},
  pages={33624--33655},
  year={2023}
}

@article{seitzer2022pitfalls,
  title={On the pitfalls of heteroscedastic uncertainty estimation with probabilistic neural networks},
  author={Seitzer, Maximilian and Tavakoli, Arash and Antic, Dimitrije and Martius, Georg},
  journal={arXiv preprint arXiv:2203.09168},
  year={2022}
}

@inproceedings{stirn2023faithful,
  title={Faithful heteroscedastic regression with neural networks},
  author={Stirn, Andrew and Wessels, Harm and Schertzer, Megan and Pereira, Laura and Sanjana, Neville and Knowles, David},
  booktitle={International Conference on Artificial Intelligence and Statistics},
  pages={5593--5613},
  year={2023},
  organization={PMLR}
}

@article{fadnavis2026leverage,
  title={Leverage-Weighted Conformal Prediction},
  author={Fadnavis, Shreyas},
  journal={arXiv preprint arXiv:2602.12693},
  year={2026}
}

@inproceedings{rossellini2024integrating,
  title={Integrating uncertainty awareness into conformalized quantile regression},
  author={Rossellini, Raphael and Barber, Rina Foygel and Willett, Rebecca},
  booktitle={International Conference on Artificial Intelligence and Statistics},
  pages={1540--1548},
  year={2024},
  organization={PMLR}
}

@article{cabezas2025epistemic,
  title={Epistemic uncertainty in conformal scores: A unified approach},
  author={Cabezas, Luben and Santos, Vagner S and Ramos, Thiago R and Izbicki, Rafael},
  journal={arXiv preprint arXiv:2502.06995},
  year={2025}
}

@article{sale2025aleatoric,
  title={Aleatoric and Epistemic Uncertainty in Conformal Prediction},
  author={Sale, Yusuf and Javanmardi, Alireza and H{\"u}llermeier, Eyke},
  year={2025},
  publisher={PMLR}
}
\bibliographystyle{tmlr}

\appendix

\section{Proofs}
\label{app:proofs}

\subsection{Proof of Proposition~\ref{prop:finite_sample_validity}}
\label{app:proof_finite_sample_validity}

\begin{proof}
Let
\[
D_{\mathrm{tr}}=\{(X_i,Y_i)\}_{i=1}^{n_{\mathrm{tr}}}
\]
denote the training set and let
\[
D_{\mathrm{cal}}
=
\{(X_i,Y_i)\}_{i=n_{\mathrm{tr}}+1}^{n_{\mathrm{tr}}+n_{\mathrm{cal}}}
\]
denote the calibration set. Write \(n=n_{\mathrm{tr}}+n_{\mathrm{cal}}\). After training, the functions
\(\mu(\cdot)\), \(h^2(\cdot)\), \(\phi(\cdot)\), and the last-layer covariance matrix \(\Sigma\) are fixed as functions of \(D_{\mathrm{tr}}\). Hence the predictive scale
\[
v(x)=h^2(x)+\phi(x)^\top\Sigma\phi(x)
\]
and the nonconformity score
\[
A(x,y)=\frac{|y-\mu(x)|}{\sqrt{v(x)}}
\]
are also fixed conditional on \(D_{\mathrm{tr}}\).

For each calibration example, define
\[
A_i=A(X_i,Y_i),
\qquad
i=n_{\mathrm{tr}}+1,\ldots,n.
\]
For the test point \((X_{n+1},Y_{n+1})\), define
\[
A_{n+1}=A(X_{n+1},Y_{n+1}).
\]
By assumption, the calibration examples and the test point are exchangeable conditional on \(D_{\mathrm{tr}}\). Since \(A(\cdot,\cdot)\) is fixed conditional on \(D_{\mathrm{tr}}\), the transformed scores
\[
A_{n_{\mathrm{tr}}+1},\ldots,A_n,A_{n+1}
\]
are also exchangeable conditional on \(D_{\mathrm{tr}}\).

Let
\[
k=\left\lceil (n_{\mathrm{cal}}+1)(1-\alpha)\right\rceil .
\]
The conformal quantile \(q_\alpha\) is the \(k\)-th order statistic of the calibration scores, with the usual conservative convention \(q_\alpha=\infty\) if \(k>n_{\mathrm{cal}}\). By the standard split conformal rank argument,
\[
\mathbb{P}\{A_{n+1}\le q_\alpha \mid D_{\mathrm{tr}}\}
\ge
1-\alpha .
\]
This argument relies only on the exchangeability of the calibration and test scores, and does not require the score distribution to be continuous.

Now observe that
\[
A_{n+1}\le q_\alpha
\]
is equivalent to
\[
\frac{|Y_{n+1}-\mu(X_{n+1})|}
{\sqrt{v(X_{n+1})}}
\le q_\alpha .
\]
Since \(v(x)>0\), this is equivalent to
\[
|Y_{n+1}-\mu(X_{n+1})|
\le
q_\alpha \sqrt{v(X_{n+1})}.
\]
Therefore,
\[
Y_{n+1}
\in
\left[
\mu(X_{n+1})-q_\alpha\sqrt{v(X_{n+1})},
\mu(X_{n+1})+q_\alpha\sqrt{v(X_{n+1})}
\right]
=
C_\alpha(X_{n+1}).
\]
Thus,
\[
\mathbb{P}\{Y_{n+1}\in C_\alpha(X_{n+1})\mid D_{\mathrm{tr}}\}
\ge
1-\alpha .
\]
Taking expectation over \(D_{\mathrm{tr}}\) gives
\[
\mathbb{P}\{Y_{n+1}\in C_\alpha(X_{n+1})\}
\ge
1-\alpha .
\]
\end{proof}

\subsection{Proof of Proposition~\ref{prop:variance_decomposition}}
\label{app:proof_variance_decomposition}

\begin{proof}
After training, the representation \(\phi(\cdot)\), the heteroscedastic variance function
\(h^2(\cdot)\), and the last-layer posterior covariance \(\Sigma\) are fixed conditional on
\(D_{\mathrm{tr}}\). Under the last-layer Gaussian approximation,
\[
w \mid D_{\mathrm{tr}} \approx \mathcal{N}(\widehat{w},\Sigma),
\]
and the conditional response model is
\[
Y \mid x,w \sim \mathcal{N}(\phi(x)^\top w,h^2(x)).
\]

By the law of total variance,
\[
\operatorname{Var}(Y\mid x,D_{\mathrm{tr}})
=
\mathbb{E}\!\left[
\operatorname{Var}(Y\mid x,w,D_{\mathrm{tr}})
\mid x,D_{\mathrm{tr}}
\right]
+
\operatorname{Var}\!\left(
\mathbb{E}[Y\mid x,w,D_{\mathrm{tr}}]
\mid x,D_{\mathrm{tr}}
\right).
\]
The first term is the conditional observation variance. Since \(h^2(x)\) is fixed once \(x\) and \(D_{\mathrm{tr}}\) are given,
\[
\mathbb{E}\!\left[
\operatorname{Var}(Y\mid x,w,D_{\mathrm{tr}})
\mid x,D_{\mathrm{tr}}
\right]
=
\mathbb{E}\!\left[h^2(x)\mid x,D_{\mathrm{tr}}\right]
=
h^2(x).
\]
The second term is the posterior variance of the last-layer predictive mean. Because
\[
\mathbb{E}[Y\mid x,w,D_{\mathrm{tr}}]=\phi(x)^\top w,
\]
we have
\[
\operatorname{Var}\!\left(
\mathbb{E}[Y\mid x,w,D_{\mathrm{tr}}]
\mid x,D_{\mathrm{tr}}
\right)
=
\operatorname{Var}(\phi(x)^\top w\mid D_{\mathrm{tr}}).
\]
Using \(w\mid D_{\mathrm{tr}}\approx \mathcal{N}(\widehat{w},\Sigma)\), this variance is
\[
\operatorname{Var}(\phi(x)^\top w\mid D_{\mathrm{tr}})
=
\phi(x)^\top \Sigma \phi(x).
\]
Combining the two terms gives
\[
\operatorname{Var}(Y\mid x,D_{\mathrm{tr}})
=
h^2(x)+\phi(x)^\top\Sigma\phi(x).
\]
\end{proof}

\subsection{Proof of Proposition~\ref{prop:heteroscedastic_precision}}
\label{app:proof_heteroscedastic_precision}

\begin{proof}
After training, the representation \(\phi(\cdot)\) and the heteroscedastic variance function \(h^2(\cdot)\) are fixed. For the final-layer weight vector \(w\), the heteroscedastic Gaussian negative log-likelihood, up to constants independent of
\(w\), is
\[
\ell(w)
=
\frac{1}{2}
\sum_{i=1}^{n_{\mathrm{tr}}}
\frac{(y_i-\phi(x_i)^\top w)^2}{h^2(x_i)} .
\]
With the Gaussian prior \(w\sim\mathcal{N}(0,\lambda^{-1}I)\), the corresponding negative log-posterior is
\[
L(w)
=
\frac{1}{2}
\sum_{i=1}^{n_{\mathrm{tr}}}
\frac{(y_i-\phi(x_i)^\top w)^2}{h^2(x_i)}
+
\frac{\lambda}{2}\|w\|_2^2 .
\]

The last-layer Laplace approximation uses the inverse Hessian of \(L(w)\) as the posterior covariance. For each training point,
\[
\nabla_w^2
\left[
\frac{1}{2}
\frac{(y_i-\phi(x_i)^\top w)^2}{h^2(x_i)}
\right]
=
\frac{\phi(x_i)\phi(x_i)^\top}{h^2(x_i)} .
\]
The Hessian of the prior term is
\[
\nabla_w^2 \left[\frac{\lambda}{2}\|w\|_2^2\right]
=
\lambda I .
\]
Therefore,
\[
\nabla_w^2 L(w)
=
\lambda I
+
\sum_{i=1}^{n_{\mathrm{tr}}}
\frac{\phi(x_i)\phi(x_i)^\top}{h^2(x_i)} .
\]

Equivalently, let \(H\in\mathbb{R}^{n_{\mathrm{tr}}\times d}\) be the feature matrix whose \(i\)-th row is \(\phi(x_i)^\top\), and let
\[
W
=
\operatorname{diag}
\left(
\frac{1}{h^2(x_1)},\ldots,
\frac{1}{h^2(x_{n_{\mathrm{tr}}})}
\right).
\]
Then
\[
H^\top W H
=
\sum_{i=1}^{n_{\mathrm{tr}}}
\frac{\phi(x_i)\phi(x_i)^\top}{h^2(x_i)} .
\]
Thus the last-layer posterior precision is
\[
\Sigma^{-1}
=
\nabla_w^2 L(w)
=
\lambda I + H^\top W H
=
\lambda I
+
\sum_{i=1}^{n_{\mathrm{tr}}}
\frac{\phi(x_i)\phi(x_i)^\top}{h^2(x_i)} .
\]
\end{proof}

\subsection{Proof of Proposition~\ref{prop:graceful_reduction}}
\label{app:proof_graceful_reduction}

\begin{proof}
Fix an input \(x\) with \(h^2(x)>0\), and consider a sequence of CLAPS predictive variances under a fixed representation and variance function,
\[
v_n(x)=h^2(x)+\phi(x)^\top\Sigma_n\phi(x).
\]
By assumption,
\[
\phi(x)^\top \Sigma_n \phi(x) \to 0.
\]
Therefore,
\[
v_n(x)
=
h^2(x)+\phi(x)^\top \Sigma_n \phi(x)
\to
h^2(x).
\]
Since \(h^2(x)>0\), we can divide by \(h^2(x)\) and obtain
\[
\frac{v_n(x)}{h^2(x)}
=
1+
\frac{\phi(x)^\top \Sigma_n \phi(x)}{h^2(x)}
\to
1.
\]
Taking square roots gives
\[
\frac{\sqrt{v_n(x)}}{h(x)}
=
\sqrt{
1+
\frac{\phi(x)^\top \Sigma_n \phi(x)}{h^2(x)}
}
\to
1.
\]
This proves that the CLAPS scale converges to the aleatoric locally adaptive scale at \(x\).

It remains to verify the stated sufficient condition. Suppose
\[
\lambda_{\min}\!\left(\lambda I+H_n^\top W_n H_n\right)\to\infty
\]
and \(\|\phi(x)\|_2<\infty\). Since
\[
\Sigma_n
=
\left(\lambda I+H_n^\top W_n H_n\right)^{-1},
\]
we have
\[
0
\le
\phi(x)^\top \Sigma_n \phi(x)
\le
\|\phi(x)\|_2^2 \lambda_{\max}(\Sigma_n).
\]
Because \(\Sigma_n\) is the inverse of \(\lambda I+H_n^\top W_n H_n\),
\[
\lambda_{\max}(\Sigma_n)
=
\frac{1}
{\lambda_{\min}\!\left(\lambda I+H_n^\top W_n H_n\right)}.
\]
Hence
\[
0
\le
\phi(x)^\top \Sigma_n \phi(x)
\le
\frac{\|\phi(x)\|_2^2}
{\lambda_{\min}\!\left(\lambda I+H_n^\top W_n H_n\right)}
\to
0.
\]
\end{proof}

\section{Additional Experimental Results}
\label{app:additional_experiments}

\subsection{Paired Comparisons for Experiment 2}
\label{app:exp2_paired}

Table~\ref{tab:exp2_paired} supplements the aggregate results in Table~\ref{tab:exp2_noise_weighted_geometry} with paired bootstrap comparisons of CLAPS against LACP, Unweighted Laplace, and LWCP.

\begin{table}[ht]
\centering
\caption{Paired comparisons for Experiment~2. All differences are computed as CLAPS minus the comparison method.}
\label{tab:exp2_paired}
\small
\setlength{\tabcolsep}{5pt}
\begin{tabular}{llcc}
\toprule
Comparison method
& Metric
& Mean difference
& 95\% CI \\
\midrule
\multirow{4}{*}{LACP}
& Marginal Coverage
& $-0.0002$
& $[-0.0007,\phantom{-}0.0003]$ \\
& Low-noise Sparse/Dense Width
& $-0.0733$
& $[-0.0907,-0.0571]$ \\
& High-noise Sparse/Dense Width
& $\phantom{-}0.0054$
& $[\phantom{-}0.0037,\phantom{-}0.0074]$ \\
& Interval Score
& $-0.0260$
& $[-0.0317,-0.0203]$ \\
\midrule
\multirow{4}{*}{Unweighted Laplace}
& Marginal Coverage
& $-0.0004$
& $[-0.0019,\phantom{-}0.0010]$ \\
& Low-noise Sparse/Dense Width
& $\phantom{-}0.9921$
& $[\phantom{-}0.8763,\phantom{-}1.1107]$ \\
& High-noise Sparse/Dense Width
& $\phantom{-}0.0131$
& $[\phantom{-}0.0100,\phantom{-}0.0163]$ \\
& Interval Score
& $\phantom{-}0.0431$
& $[\phantom{-}0.0320,\phantom{-}0.0549]$ \\
\midrule
\multirow{4}{*}{LWCP}
& Marginal Coverage
& $-0.0002$
& $[-0.0028,\phantom{-}0.0024]$ \\
& Low-noise Sparse/Dense Width
& $\phantom{-}4.9276$
& $[\phantom{-}4.5351,\phantom{-}5.3550]$ \\
& High-noise Sparse/Dense Width
& $\phantom{-}0.8738$
& $[\phantom{-}0.7657,\phantom{-}0.9803]$ \\
& Interval Score
& $-0.3402$
& $[-0.4509,-0.2360]$ \\
\bottomrule
\end{tabular}
\end{table}

\subsection{Full Results for the Training-Size Sweep}
\label{app:exp3_training_size}

Table~\ref{tab:exp3_training_size_full} reports the complete numerical results for the training-size sweep in Figure~\ref{fig:exp3_regime_study}, including marginal coverage, sparse-to-dense width ratio, sparse-region interval score, and the CLAPS epistemic fraction.

\begin{table}[ht]
\centering
\caption{Full results for the training-size sweep in Experiment~3. Values are means and standard errors over random seeds.}
\label{tab:exp3_training_size_full}
\small
\setlength{\tabcolsep}{3.5pt}
\resizebox{\textwidth}{!}{
\begin{tabular}{clcccc}
\toprule
Training size & Method & Marginal Cov. & Sparse/Dense Width & Sparse Interval Score & CLAPS Epi. Frac. \\
\midrule
\multirow{4}{*}{\(100\)}
& LACP                 & \(0.9021 \pm 0.0014\) & \(1.0725 \pm 0.0451\) & \(4.4936 \pm 0.2258\) & --- \\
& Unweighted Laplace   & \(0.9013 \pm 0.0014\) & \(1.0629 \pm 0.0383\) & \(4.3667 \pm 0.1988\) & --- \\
& LWCP                 & \(0.9009 \pm 0.0011\) & \(1.2229 \pm 0.0156\) & \(3.9478 \pm 0.1293\) & --- \\
& CLAPS                & \(0.9021 \pm 0.0014\) & \(1.0752 \pm 0.0440\) & \(4.4629 \pm 0.2194\) & \(0.0301 \pm 0.0090\) \\
\midrule
\multirow{4}{*}{\(250\)}
& LACP                 & \(0.8984 \pm 0.0014\) & \(1.3236 \pm 0.0914\) & \(5.3503 \pm 0.2354\) & --- \\
& Unweighted Laplace   & \(0.8984 \pm 0.0015\) & \(1.2862 \pm 0.0778\) & \(5.2994 \pm 0.2291\) & --- \\
& LWCP                 & \(0.9002 \pm 0.0013\) & \(1.1381 \pm 0.0089\) & \(4.2681 \pm 0.1056\) & --- \\
& CLAPS                & \(0.8984 \pm 0.0015\) & \(1.3257 \pm 0.0899\) & \(5.2977 \pm 0.2288\) & \(0.0271 \pm 0.0077\) \\
\midrule
\multirow{4}{*}{\(500\)}
& LACP                 & \(0.9014 \pm 0.0012\) & \(2.8760 \pm 0.3218\) & \(5.1054 \pm 0.1851\) & --- \\
& Unweighted Laplace   & \(0.9018 \pm 0.0012\) & \(2.6310 \pm 0.2664\) & \(5.0070 \pm 0.1862\) & --- \\
& LWCP                 & \(0.8990 \pm 0.0012\) & \(1.0831 \pm 0.0065\) & \(4.3441 \pm 0.0912\) & --- \\
& CLAPS                & \(0.9014 \pm 0.0012\) & \(2.8408 \pm 0.3158\) & \(5.0546 \pm 0.1813\) & \(0.0378 \pm 0.0075\) \\
\midrule
\multirow{4}{*}{\(1000\)}
& LACP                 & \(0.8987 \pm 0.0013\) & \(4.2805 \pm 0.1929\) & \(4.1594 \pm 0.0799\) & --- \\
& Unweighted Laplace   & \(0.8973 \pm 0.0013\) & \(4.0256 \pm 0.1795\) & \(4.1276 \pm 0.0832\) & --- \\
& LWCP                 & \(0.8967 \pm 0.0016\) & \(1.0526 \pm 0.0029\) & \(4.1745 \pm 0.0667\) & --- \\
& CLAPS                & \(0.8986 \pm 0.0014\) & \(4.2434 \pm 0.1902\) & \(4.1176 \pm 0.0812\) & \(0.0391 \pm 0.0045\) \\
\midrule
\multirow{4}{*}{\(2000\)}
& LACP                 & \(0.9015 \pm 0.0009\) & \(4.6009 \pm 0.1920\) & \(3.6394 \pm 0.0941\) & --- \\
& Unweighted Laplace   & \(0.9016 \pm 0.0010\) & \(4.3895 \pm 0.1797\) & \(3.5955 \pm 0.0952\) & --- \\
& LWCP                 & \(0.9026 \pm 0.0012\) & \(1.0269 \pm 0.0011\) & \(3.8124 \pm 0.1144\) & --- \\
& CLAPS                & \(0.9016 \pm 0.0008\) & \(4.5828 \pm 0.1904\) & \(3.6051 \pm 0.0939\) & \(0.0243 \pm 0.0021\) \\
\bottomrule
\end{tabular}
}
\end{table}

\subsection{Full Results for the Local-Support Sweep}
\label{app:exp3_local_support}

Table~\ref{tab:exp3_local_support_full} reports the complete numerical results for the local-support sweep in Figure~\ref{fig:exp3_regime_study}, including marginal coverage, sparse-to-dense width ratio, sparse-region interval score, and the CLAPS epistemic fraction.

\begin{table}[ht]
\centering
\caption{Full results for the local-support sweep in Experiment~3. Values are means and standard errors over random seeds.}
\label{tab:exp3_local_support_full}
\small
\setlength{\tabcolsep}{3.5pt}
\resizebox{\textwidth}{!}{
\begin{tabular}{clcccc}
\toprule
Sparse retention & Method & Marginal Cov. & Sparse/Dense Width & Sparse Interval Score & CLAPS Epi. Frac. \\
\midrule
\multirow{4}{*}{\(0.01\)}
& LACP               & \(0.8979 \pm 0.0014\) & \(3.4513 \pm 0.1974\) & \(6.1947 \pm 0.2428\) & --- \\
& Unweighted Laplace & \(0.8980 \pm 0.0013\) & \(3.2555 \pm 0.1799\) & \(6.1989 \pm 0.2401\) & --- \\
& LWCP               & \(0.8986 \pm 0.0012\) & \(1.1228 \pm 0.0074\) & \(4.8564 \pm 0.1338\) & --- \\
& CLAPS              & \(0.8976 \pm 0.0013\) & \(3.4752 \pm 0.1998\) & \(6.1272 \pm 0.2418\) & \(0.0463 \pm 0.0058\) \\
\midrule
\multirow{4}{*}{\(0.025\)}
& LACP               & \(0.8991 \pm 0.0015\) & \(3.6762 \pm 0.2666\) & \(5.4421 \pm 0.2664\) & --- \\
& Unweighted Laplace & \(0.8988 \pm 0.0015\) & \(3.4596 \pm 0.2435\) & \(5.4045 \pm 0.2603\) & --- \\
& LWCP               & \(0.8980 \pm 0.0013\) & \(1.0980 \pm 0.0060\) & \(4.6070 \pm 0.1404\) & --- \\
& CLAPS              & \(0.8995 \pm 0.0015\) & \(3.6734 \pm 0.2658\) & \(5.3862 \pm 0.2580\) & \(0.0389 \pm 0.0050\) \\
\midrule
\multirow{4}{*}{\(0.05\)}
& LACP               & \(0.8992 \pm 0.0016\) & \(3.8610 \pm 0.2456\) & \(4.9855 \pm 0.2205\) & --- \\
& Unweighted Laplace & \(0.8996 \pm 0.0015\) & \(3.6350 \pm 0.2202\) & \(4.9527 \pm 0.2222\) & --- \\
& LWCP               & \(0.8985 \pm 0.0014\) & \(1.0671 \pm 0.0034\) & \(4.4696 \pm 0.1413\) & --- \\
& CLAPS              & \(0.8993 \pm 0.0016\) & \(3.8468 \pm 0.2434\) & \(4.9216 \pm 0.2184\) & \(0.0484 \pm 0.0045\) \\
\midrule
\multirow{4}{*}{\(0.10\)}
& LACP               & \(0.8987 \pm 0.0013\) & \(4.2805 \pm 0.1929\) & \(4.1594 \pm 0.0799\) & --- \\
& Unweighted Laplace & \(0.8973 \pm 0.0013\) & \(4.0256 \pm 0.1795\) & \(4.1276 \pm 0.0832\) & --- \\
& LWCP               & \(0.8967 \pm 0.0016\) & \(1.0526 \pm 0.0029\) & \(4.1745 \pm 0.0667\) & --- \\
& CLAPS              & \(0.8986 \pm 0.0014\) & \(4.2434 \pm 0.1902\) & \(4.1176 \pm 0.0812\) & \(0.0391 \pm 0.0045\) \\
\midrule
\multirow{4}{*}{\(0.20\)}
& LACP               & \(0.9001 \pm 0.0016\) & \(4.1152 \pm 0.2370\) & \(3.8664 \pm 0.0585\) & --- \\
& Unweighted Laplace & \(0.8995 \pm 0.0014\) & \(3.8626 \pm 0.2165\) & \(3.8101 \pm 0.0626\) & --- \\
& LWCP               & \(0.9008 \pm 0.0013\) & \(1.0305 \pm 0.0018\) & \(4.1577 \pm 0.1221\) & --- \\
& CLAPS              & \(0.9001 \pm 0.0015\) & \(4.0739 \pm 0.2335\) & \(3.8247 \pm 0.0577\) & \(0.0353 \pm 0.0044\) \\
\midrule
\multirow{4}{*}{\(1.00\)}
& LACP               & \(0.8987 \pm 0.0014\) & \(1.9473 \pm 0.1824\) & \(2.7546 \pm 0.0748\) & --- \\
& Unweighted Laplace & \(0.8996 \pm 0.0013\) & \(1.8399 \pm 0.1580\) & \(2.7015 \pm 0.0811\) & --- \\
& LWCP               & \(0.8996 \pm 0.0012\) & \(0.9949 \pm 0.0009\) & \(2.7107 \pm 0.0971\) & --- \\
& CLAPS              & \(0.8990 \pm 0.0014\) & \(1.9259 \pm 0.1779\) & \(2.7362 \pm 0.0755\) & \(0.0200 \pm 0.0039\) \\
\bottomrule
\end{tabular}
}
\end{table}

\subsection{Raw-Input Support-Stratified Results}
\label{app:exp4_raw_support}

Table~\ref{tab:exp4_raw_support} reports the CLAPS--LACP differences across support quintiles defined by nearest-neighbor distances in the standardized raw-input space. The results complement the learned-feature stratification reported in Table~\ref{tab:exp4_feature_quintiles}.

\begin{table}[ht]
\centering
\caption{Raw-input support-stratified comparison between CLAPS and LACP in Experiment~4. All differences are computed as CLAPS minus LACP and reported as means and standard errors. Q1 and Q5 denote the strongest and weakest support, respectively.}
\label{tab:exp4_raw_support}
\small
\setlength{\tabcolsep}{6pt}
\begin{tabular}{lccc}
\toprule
Quintile & \(\Delta\) Coverage & \(\Delta\) Width & \(\Delta\) Score \\
\midrule
Q1
& \(-0.0040 \pm 0.0011\)
& \(-2.21 \pm 0.30\%\)
& \(-1.02 \pm 0.26\%\) \\
Q2
& \(-0.0041 \pm 0.0017\)
& \(-1.89 \pm 0.27\%\)
& \(-0.95 \pm 0.21\%\) \\
Q3
& \(-0.0024 \pm 0.0012\)
& \(-1.63 \pm 0.26\%\)
& \(-1.05 \pm 0.24\%\) \\
Q4
& \(-0.0034 \pm 0.0015\)
& \(-0.99 \pm 0.25\%\)
& \(-0.57 \pm 0.23\%\) \\
Q5
& \(\phantom{-}0.0065 \pm 0.0020\)
& \(\phantom{-}0.73 \pm 0.34\%\)
& \(-1.23 \pm 0.47\%\) \\
\bottomrule
\end{tabular}
\end{table}

\section{Sensitivity Analyses}
\label{app:sensitivity_analyses}

\subsection{Common Experimental Design}
\label{app:sensitivity_common_design}

The sensitivity analyses follow the real-data protocol of Experiment~4. We use \(20\) random seeds, a \(60/20/20\) train--calibration--test split, and at most \(5{,}000\) observations per dataset. The prior-precision analysis uses all eight real-data benchmarks, while the representation-dimension and variance-floor analyses use Concrete, Energy, Bike, and California. Data preprocessing, outer splits, inner training splits, and random initialization are fixed across the configurations compared within each analysis. These analyses use independent runs under the same outer protocol, so their aggregate values may differ slightly from the main benchmark.

The OOD stress test follows the ten-dimensional synthetic design of Experiment~1, with \(30\) random seeds, \(1{,}000\) training observations, and \(2{,}000\) calibration observations per seed. The fitted regressor, last-layer covariance, and conformal quantiles are held fixed across the in-distribution and shifted test conditions. OOD observations are used only for evaluation, without model refitting or recalibration.

Each configuration fits a heteroscedastic neural regressor that produces a predictive mean, an aleatoric variance \(h^2(x)\), and a learned representation \(\phi(x)\). Within each fitted model, LACP and CLAPS share these quantities. LACP uses the normalization scale
\[
h(x),
\]
and CLAPS uses
\[
\sqrt{h^2(x)+\phi(x)^\top\Sigma_\lambda\phi(x)},
\qquad
\Sigma_\lambda=\left(\lambda I + H^\top W H\right)^{-1}.
\]
The OOD analysis additionally includes an unweighted last-layer variant constructed from the same representation. The regressor is retrained when the representation dimension or variance floor changes because these settings affect the learned feature map and variance head. Fixed prior-precision configurations reuse the same fitted regressor and vary only the last-layer covariance.

Unless fixed explicitly, the prior precision is selected independently for each dataset and seed from \(\{10^{-4},10^{-3},10^{-2},10^{-1},1,10\}\). The training split is divided into covariance-fitting and validation subsets, and the candidate minimizing Gaussian predictive negative log-likelihood on the validation subset is selected. The posterior covariance is then reconstructed from the complete training split. Calibration and test responses are excluded from prior-precision selection.

We report marginal coverage and relative changes in average width and interval score against matched LACP. Support is measured independently by the average distance to the \(20\) nearest training representations, with Q5 denoting the weakest-support quintile. The OOD analysis also measures the corresponding distance in standardized input space. The epistemic fraction is defined as
\[
\frac{\phi(x)^\top\Sigma_\lambda\phi(x)}
{h^2(x)+\phi(x)^\top\Sigma_\lambda\phi(x)}.
\]
Tables report means and standard errors over dataset--seed or synthetic-seed runs, as appropriate. Paired bootstrap intervals account for variation across seeds, with hierarchical resampling across datasets and seeds for the real-data analyses.

\subsection{Prior Precision Selection}
\label{app:prior_precision}

\paragraph{Experimental setup.}
We evaluate the fixed grid \(\lambda\in\{10^{-5},10^{-4},10^{-3},10^{-2},10^{-1},1,10,100\}\) using a \(64\)-dimensional representation. A single fitted heteroscedastic regressor is shared across all fixed-\(\lambda\) configurations for each dataset and seed. We also evaluate the training-only selection procedure used in the main experiments.

\paragraph{Results and analysis.}
Table~\ref{tab:prior_precision} shows that increasing the prior precision smoothly contracts the posterior correction. The epistemic contribution and the differences between CLAPS and LACP decrease together, while marginal coverage remains stable. The epistemic fraction is consistently larger in Q5 than over the complete test set, and the Q5 interval score improves throughout the fixed grid. Training-only selection preserves this support-dependent pattern and improves both overall and weak-support interval scores.

\begin{table}[ht]
    \centering
    \caption{Prior-precision selection and sensitivity on the eight real-data benchmarks. Panel~(a) reports means and standard errors over dataset--seed runs. Changes in width and interval score are relative to matched LACP. Panel~(b) reports hierarchical paired bootstrap comparisons between training-only selected CLAPS and LACP.}
    \label{tab:prior_precision}
    \small
    \setlength{\tabcolsep}{3.5pt}

    (a) Fixed prior-precision sensitivity and training-only selection.

    \vspace{0.4em}

    \resizebox{\linewidth}{!}{
    \begin{tabular}{lccccccc}
        \toprule
        \multirow[c]{2}{*}{\(\lambda\)}
        & \multirow[c]{2}{*}{Coverage}
        & \multicolumn{2}{c}{Overall}
        & \multicolumn{2}{c}{Weak-Q5}
        & \multicolumn{2}{c}{Epistemic fraction} \\
        \cmidrule(lr){3-4}\cmidrule(lr){5-6}\cmidrule(lr){7-8}
        & & \(\Delta\) Width & \(\Delta\) Score & \(\Delta\) Width & \(\Delta\) Score & Overall & Weak-Q5 \\
        \midrule
        \(10^{-5}\) & \(0.9014 \pm 0.0023\) & \(-2.10 \pm 0.41\%\) & \(-2.20 \pm 0.32\%\) & \( 1.45 \pm 0.66\%\) & \(-3.68 \pm 0.56\%\) & \( 9.93 \pm 0.83\%\) & \(16.07 \pm 0.87\%\) \\
        \(10^{-4}\) & \(0.9015 \pm 0.0023\) & \(-2.11 \pm 0.41\%\) & \(-2.19 \pm 0.32\%\) & \( 1.41 \pm 0.66\%\) & \(-3.67 \pm 0.56\%\) & \( 9.87 \pm 0.83\%\) & \(15.96 \pm 0.88\%\) \\
        \(10^{-3}\) & \(0.9017 \pm 0.0023\) & \(-2.09 \pm 0.41\%\) & \(-2.15 \pm 0.32\%\) & \( 1.29 \pm 0.65\%\) & \(-3.64 \pm 0.56\%\) & \( 9.67 \pm 0.82\%\) & \(15.60 \pm 0.87\%\) \\
        \(10^{-2}\) & \(0.9016 \pm 0.0023\) & \(-2.03 \pm 0.41\%\) & \(-2.01 \pm 0.31\%\) & \( 0.89 \pm 0.62\%\) & \(-3.46 \pm 0.52\%\) & \( 8.95 \pm 0.78\%\) & \(14.34 \pm 0.82\%\) \\
        \(10^{-1}\) & \(0.9021 \pm 0.0023\) & \(-1.85 \pm 0.37\%\) & \(-1.70 \pm 0.27\%\) & \( 0.20 \pm 0.51\%\) & \(-2.92 \pm 0.43\%\) & \( 7.47 \pm 0.69\%\) & \(11.78 \pm 0.71\%\) \\
        \(1\) & \(0.9033 \pm 0.0023\) & \(-1.42 \pm 0.30\%\) & \(-1.27 \pm 0.23\%\) & \(-0.24 \pm 0.38\%\) & \(-2.11 \pm 0.32\%\) & \( 5.61 \pm 0.56\%\) & \( 8.62 \pm 0.57\%\) \\
        \(10\) & \(0.9032 \pm 0.0023\) & \(-1.11 \pm 0.24\%\) & \(-0.90 \pm 0.18\%\) & \(-0.52 \pm 0.28\%\) & \(-1.42 \pm 0.24\%\) & \( 3.84 \pm 0.42\%\) & \( 5.67 \pm 0.42\%\) \\
        \(100\) & \(0.9033 \pm 0.0022\) & \(-0.81 \pm 0.18\%\) & \(-0.58 \pm 0.13\%\) & \(-0.57 \pm 0.20\%\) & \(-0.86 \pm 0.17\%\) & \( 2.41 \pm 0.28\%\) & \( 3.40 \pm 0.29\%\) \\
        \midrule
        Training-only selected & \(0.9040 \pm 0.0023\) & \(-1.13 \pm 0.25\%\) & \(-1.14 \pm 0.20\%\) & \( 0.24 \pm 0.37\%\) & \(-2.02 \pm 0.32\%\) & \( 4.89 \pm 0.44\%\) & \( 7.90 \pm 0.53\%\) \\
        \bottomrule
    \end{tabular}
    }

    \vspace{1em}

    (b) Hierarchical paired bootstrap for selected CLAPS minus LACP.

    \vspace{0.4em}

    \resizebox{0.88\linewidth}{!}{
    \begin{tabular}{lcccc}
        \toprule
        \multirow[c]{2}{*}{Metric}
        & \multicolumn{2}{c}{Overall}
        & \multicolumn{2}{c}{Weak-Q5} \\
        \cmidrule(lr){2-3}\cmidrule(lr){4-5}
        & Estimate & \(95\%\) CI & Estimate & \(95\%\) CI \\
        \midrule
        \(\Delta\) Coverage & \(0.0007\) & \([-0.0012,\ 0.0026]\) & \(0.0084\) & \([0.0034,\ 0.0146]\) \\
        \(\Delta\) Width & \(-1.1330\%\) & \([-2.9011\%,\ -0.0583\%]\) & \(0.2377\%\) & \([-2.2704\%,\ 2.0608\%]\) \\
        \(\Delta\) Interval score & \(-1.1408\%\) & \([-2.4834\%,\ -0.2405\%]\) & \(-2.0174\%\) & \([-3.6750\%,\ -0.6623\%]\) \\
        \bottomrule
    \end{tabular}
    }
\end{table}

\subsection{Representation Dimension}
\label{app:feature_dimension}

\paragraph{Experimental setup.}
We vary the learned representation dimension over \(d\in\{16,32,64,128\}\), while fixing the hidden-layer width at \(64\). Because the representation dimension changes the learned feature map and variance head, the heteroscedastic regressor is retrained for every dataset--seed--dimension combination. We additionally record model size, training cost, prior-selection cost, covariance construction, the additional quadratic-form cost after feature extraction, and covariance memory.

\paragraph{Results and analysis.}
Table~\ref{tab:representation_dimension} shows stable marginal coverage and improved interval scores across the evaluated dimensions. Larger representations produce stronger epistemic contributions, with the increase concentrated in Q5, and are associated with larger weak-support improvements. Posterior computation and memory increase with dimension but remain small relative to neural-network training and prediction.

\begin{table}[ht]
    \centering
    \caption{Sensitivity to the learned representation dimension. Panel~(a) reports predictive performance and posterior contribution. Changes in width and interval score are relative to matched LACP. Panel~(b) reports model and last-layer posterior costs.}
    \label{tab:representation_dimension}
    \small
    \setlength{\tabcolsep}{3.2pt}

    (a) Predictive performance and posterior contribution.

    \vspace{0.4em}

    \resizebox{\linewidth}{!}{
    \begin{tabular}{ccccccccc}
        \toprule
        \multirow[c]{2}{*}{\shortstack{Feature\\dimension}}
        & \multirow[c]{2}{*}{Coverage}
        & \multicolumn{2}{c}{Overall}
        & \multicolumn{3}{c}{Weak-Q5}
        & \multicolumn{2}{c}{Epistemic fraction} \\
        \cmidrule(lr){3-4}\cmidrule(lr){5-7}\cmidrule(lr){8-9}
        & & \(\Delta\) Width & \(\Delta\) Score & \(\Delta\) Coverage & \(\Delta\) Width & \(\Delta\) Score & Overall & Weak-Q5 \\
        \midrule
        \(16\) & \(0.8992 \pm 0.0029\) & \(-0.36 \pm 0.12\%\) & \(-0.33 \pm 0.10\%\) & \(0.0019 \pm 0.0010\) & \( 0.26 \pm 0.15\%\) & \(-0.90 \pm 0.25\%\) & \(1.64 \pm 0.14\%\) & \( 3.33 \pm 0.26\%\) \\
        \(32\) & \(0.9000 \pm 0.0029\) & \(-1.09 \pm 0.22\%\) & \(-0.60 \pm 0.16\%\) & \(0.0010 \pm 0.0011\) & \(-0.12 \pm 0.25\%\) & \(-1.39 \pm 0.34\%\) & \(2.59 \pm 0.22\%\) & \( 5.35 \pm 0.42\%\) \\
        \(64\) & \(0.8996 \pm 0.0025\) & \(-0.63 \pm 0.22\%\) & \(-0.82 \pm 0.23\%\) & \(0.0091 \pm 0.0020\) & \( 1.11 \pm 0.44\%\) & \(-1.92 \pm 0.53\%\) & \(3.96 \pm 0.37\%\) & \( 8.09 \pm 0.71\%\) \\
        \(128\) & \(0.9016 \pm 0.0023\) & \(-1.18 \pm 0.26\%\) & \(-1.15 \pm 0.28\%\) & \(0.0125 \pm 0.0029\) & \( 1.51 \pm 0.55\%\) & \(-2.97 \pm 0.71\%\) & \(4.78 \pm 0.43\%\) & \(10.36 \pm 0.82\%\) \\
        \bottomrule
    \end{tabular}
    }

    \vspace{1em}

    (b) Computational cost.

    \vspace{0.4em}

    \resizebox{\linewidth}{!}{
    \begin{tabular}{ccccccc}
        \toprule
        \multirow[c]{2}{*}{\shortstack{Feature\\dimension}}
        & \multicolumn{2}{c}{Neural model}
        & \multicolumn{4}{c}{Last-layer posterior} \\
        \cmidrule(lr){2-3}\cmidrule(lr){4-7}
        & Parameters & Training time & Selection time & Covariance time & Quadratic-form time & Memory \\
        \midrule
        \(16\) & \(5{,}874 \pm 12\) & \(0.62 \pm 0.02\) s & \(0.003 \pm 0.000\) s & \(0.287 \pm 0.014\) ms & \(0.157 \pm 0.007\) \(\mu\)s & \(0.0022 \pm 0.0000\) MB \\
        \(32\) & \(6{,}946 \pm 12\) & \(0.57 \pm 0.01\) s & \(0.008 \pm 0.002\) s & \(1.237 \pm 0.451\) ms & \(0.467 \pm 0.135\) \(\mu\)s & \(0.0083 \pm 0.0000\) MB \\
        \(64\) & \(9{,}090 \pm 12\) & \(0.56 \pm 0.02\) s & \(0.024 \pm 0.004\) s & \(2.745 \pm 0.543\) ms & \(0.737 \pm 0.132\) \(\mu\)s & \(0.0322 \pm 0.0000\) MB \\
        \(128\) & \(13{,}378 \pm 12\) & \(0.56 \pm 0.02\) s & \(0.055 \pm 0.009\) s & \(7.285 \pm 1.500\) ms & \(1.539 \pm 0.116\) \(\mu\)s & \(0.1270 \pm 0.0000\) MB \\
        \bottomrule
    \end{tabular}
    }
\end{table}

\subsection{Variance Floor}
\label{app:variance_floor}

\paragraph{Experimental setup.}
We vary the variance floor over \(\epsilon\in\{10^{-6},10^{-5},10^{-4},10^{-3}\}\) with the representation dimension fixed at \(64\). Because the floor enters the variance parameterization and training likelihood, the heteroscedastic regressor is retrained for each dataset--seed--floor combination. We compare each result with matched LACP and directly with the default floor \(\epsilon=10^{-4}\). The numerical diagnostics measure floor activity, the upper tail of the precision weights \(1/h^2(x)\), and the condition of the last-layer system.

\paragraph{Results and analysis.}
Table~\ref{tab:variance_floor} shows stable predictive performance and support-dependent posterior behavior throughout the floor grid. Increasing the floor reduces extreme inverse-variance weights while leaving the epistemic fractions and last-layer condition broadly unchanged. Direct comparisons with the default floor show no systematic shift in coverage, width, or interval score.

\begin{table}[ht]
    \centering
    \caption{Sensitivity to the variance floor. Panel~(a) reports predictive performance and posterior contribution. Changes in width and interval score are relative to matched LACP. Panel~(b) reports variance-head and last-layer numerical diagnostics. Panel~(c) reports hierarchical paired bootstrap comparisons with the default floor \(\epsilon=10^{-4}\).}
    \label{tab:variance_floor}
    \small
    \setlength{\tabcolsep}{3.1pt}

    (a) Predictive performance and posterior contribution.

    \vspace{0.4em}

    \resizebox{\linewidth}{!}{
    \begin{tabular}{ccccccccc}
        \toprule
        \multirow[c]{2}{*}{\shortstack{Variance\\floor}}
        & \multirow[c]{2}{*}{Coverage}
        & \multicolumn{2}{c}{Overall}
        & \multicolumn{3}{c}{Weak-Q5}
        & \multicolumn{2}{c}{Epistemic fraction} \\
        \cmidrule(lr){3-4}\cmidrule(lr){5-7}\cmidrule(lr){8-9}
        & & \(\Delta\) Width & \(\Delta\) Score & \(\Delta\) Coverage & \(\Delta\) Width & \(\Delta\) Score & Overall & Weak-Q5 \\
        \midrule
        \(10^{-6}\) & \(0.9017 \pm 0.0025\) & \(-0.78 \pm 0.27\%\) & \(-0.87 \pm 0.17\%\) & \(0.0117 \pm 0.0025\) & \(1.04 \pm 0.41\%\) & \(-1.93 \pm 0.42\%\) & \(4.03 \pm 0.39\%\) & \(8.17 \pm 0.73\%\) \\
        \(10^{-5}\) & \(0.9018 \pm 0.0025\) & \(-0.78 \pm 0.26\%\) & \(-0.89 \pm 0.17\%\) & \(0.0130 \pm 0.0024\) & \(1.08 \pm 0.42\%\) & \(-2.04 \pm 0.45\%\) & \(4.08 \pm 0.40\%\) & \(8.27 \pm 0.76\%\) \\
        \(10^{-4}\) & \(0.9013 \pm 0.0023\) & \(-0.89 \pm 0.26\%\) & \(-0.82 \pm 0.17\%\) & \(0.0104 \pm 0.0024\) & \(0.92 \pm 0.41\%\) & \(-1.98 \pm 0.44\%\) & \(4.03 \pm 0.40\%\) & \(8.17 \pm 0.75\%\) \\
        \(10^{-3}\) & \(0.9013 \pm 0.0024\) & \(-0.64 \pm 0.24\%\) & \(-0.81 \pm 0.16\%\) & \(0.0090 \pm 0.0020\) & \(1.13 \pm 0.39\%\) & \(-1.96 \pm 0.43\%\) & \(3.94 \pm 0.38\%\) & \(7.95 \pm 0.71\%\) \\
        \bottomrule
    \end{tabular}
    }

    \vspace{1em}

    (b) Variance-head and last-layer numerical diagnostics.

    \vspace{0.4em}

    \resizebox{\linewidth}{!}{
    \begin{tabular}{cccccc}
        \toprule
        \multirow[c]{2}{*}{\shortstack{Variance\\floor}}
        & \multicolumn{2}{c}{Floor activity}
        & \multicolumn{2}{c}{Precision weight \(1/h^2\)}
        & \multicolumn{1}{c}{Last-layer system} \\
        \cmidrule(lr){2-3}\cmidrule(lr){4-5}\cmidrule(lr){6-6}
        & Mean contribution & Dominant rate & \(99\)th percentile & Maximum & \(\log_{10}\) condition number \\
        \midrule
        \(10^{-6}\) & \(0.008 \pm 0.001\%\) & \(0.000 \pm 0.000\%\) & \(686.8 \pm 97.8\) & \(2008.9 \pm 285.6\) & \(5.818 \pm 0.161\) \\
        \(10^{-5}\) & \(0.076 \pm 0.011\%\) & \(0.001 \pm 0.001\%\) & \(694.0 \pm 96.2\) & \(1960.6 \pm 272.8\) & \(5.836 \pm 0.167\) \\
        \(10^{-4}\) & \(0.730 \pm 0.105\%\) & \(0.003 \pm 0.002\%\) & \(659.1 \pm 87.2\) & \(1551.2 \pm 189.8\) & \(5.820 \pm 0.163\) \\
        \(10^{-3}\) & \(7.002 \pm 0.997\%\) & \(5.041 \pm 0.959\%\) & \(391.3 \pm 41.3\) & \(544.9 \pm 43.3\) & \(5.759 \pm 0.156\) \\
        \bottomrule
    \end{tabular}
    }

    \vspace{1em}

    (c) Hierarchical paired bootstrap relative to the default floor \(10^{-4}\).

    \vspace{0.4em}

    \resizebox{\linewidth}{!}{
    \begin{tabular}{ccccccc}
        \toprule
        \multirow[c]{2}{*}{\shortstack{Variance\\floor}}
        & \multicolumn{2}{c}{\(\Delta\) Coverage}
        & \multicolumn{2}{c}{\(\Delta\) Width}
        & \multicolumn{2}{c}{\(\Delta\) Interval score} \\
        \cmidrule(lr){2-3}\cmidrule(lr){4-5}\cmidrule(lr){6-7}
        & Estimate & \(95\%\) CI & Estimate & \(95\%\) CI & Estimate & \(95\%\) CI \\
        \midrule
        \(10^{-6}\) & \(0.0004\) & \([-0.0010,\ 0.0020]\) & \(0.0378\%\) & \([-0.4536\%,\ 0.5227\%]\) & \(-0.0584\%\) & \([-0.4272\%,\ 0.3265\%]\) \\
        \(10^{-5}\) & \(0.0005\) & \([-0.0018,\ 0.0044]\) & \(0.1557\%\) & \([-0.4147\%,\ 0.9553\%]\) & \(-0.0190\%\) & \([-0.4749\%,\ 0.3736\%]\) \\
        \(10^{-3}\) & \(0.0001\) & \([-0.0023,\ 0.0024]\) & \(-1.0597\%\) & \([-3.1793\%,\ 0.2353\%]\) & \(-0.8179\%\) & \([-2.1218\%,\ 0.1731\%]\) \\
        \bottomrule
    \end{tabular}
    }
\end{table}

\subsection{OOD Stress Test under a Fixed Representation}
\label{app:ood_stress}

\paragraph{Experimental setup.}
We consider three test conditions under the synthetic design of Experiment~1. The ID reference follows the original target distribution. Feature-visible OOD inputs are distant from the training sample in both standardized input space and learned-feature space. Feature-collapsed OOD inputs exhibit substantial raw-input and conditional-mean shifts while remaining within the upper range of the ID feature-distance distribution. We compare LACP, Unweighted Laplace, and CLAPS using input- and feature-space distances, predictive error, epistemic contribution, interval width, coverage, and interval score. Coverage under the shifted conditions is reported as an empirical stress-test measure because the OOD samples are not exchangeable with the calibration sample.

\begin{figure}[ht]
    \centering
    \includegraphics[width=0.82\linewidth]{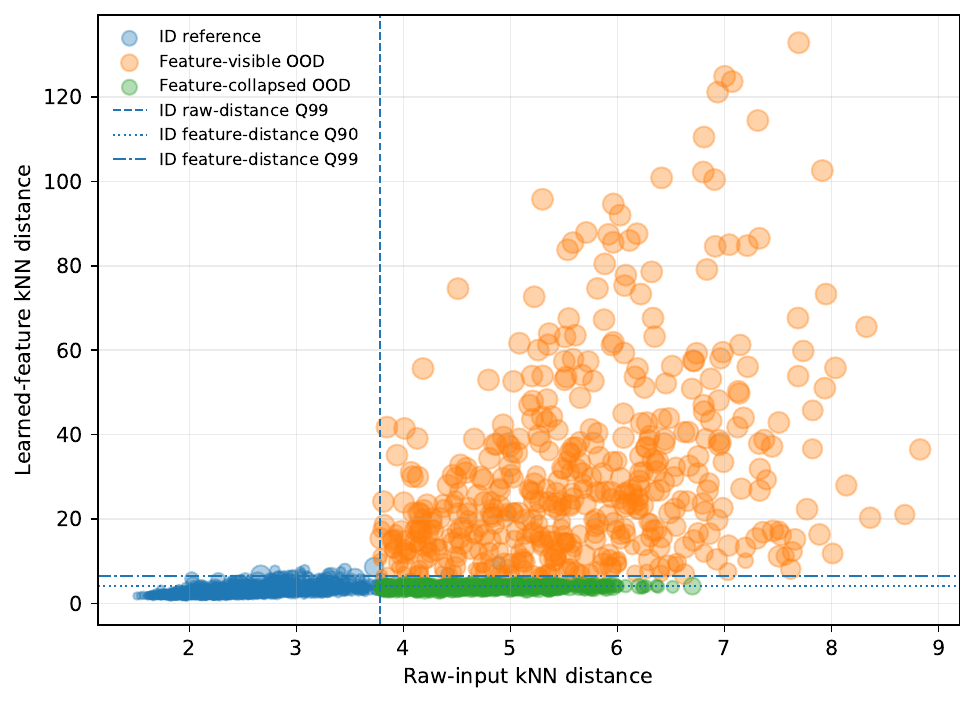}
    \caption{Raw-input and learned-feature distances under the ID reference, feature-visible OOD, and feature-collapsed OOD conditions. The dotted and dash-dotted lines denote the \(90\)th and \(99\)th percentiles of the corresponding ID distances.}
    \label{fig:ood_stress_geometry}
\end{figure}

\paragraph{Results and analysis.}
Figure~\ref{fig:ood_stress_geometry} shows that the feature-visible condition separates clearly from the ID reference in both spaces, whereas the feature-collapsed condition remains close to the ID range in learned-feature distance despite its raw-input displacement. Table~\ref{tab:ood_stress} shows a correspondingly strong CLAPS correction under feature-visible shift, with increased epistemic contribution and interval expansion that improve coverage and interval score relative to LACP. Under feature collapse, the conditional mean and prediction error change substantially, but the last-layer correction remains modest and the predictive performance of all methods deteriorates. The unweighted variant exhibits the same qualitative dependence on the fixed representation. These results illustrate that last-layer uncertainty responds to shifts preserved by the learned feature map and can remain limited when target-relevant changes are compressed into an ID-like representation.

\begin{table}[ht]
\centering
\caption{OOD stress test under a fixed representation. Values are means and standard errors over \(30\) random seeds. The width ratio compares CLAPS with matched LACP.}
\label{tab:ood_stress}
\small

(a) Shift and uncertainty diagnostics

\resizebox{\textwidth}{!}{
\begin{tabular}{lcccccc}
\toprule
& \multicolumn{2}{c}{kNN distance}
& \multicolumn{2}{c}{Target shift}
& \multicolumn{2}{c}{CLAPS response} \\
\cmidrule(lr){2-3}
\cmidrule(lr){4-5}
\cmidrule(lr){6-7}
Condition
& Raw input
& Learned feature
& Mean shift
& Mean RMSE
& Epi. fraction
& Width ratio \\
\midrule
ID reference
& \(2.5013 \pm 0.0034\)
& \(3.2301 \pm 0.0445\)
& \(0.0000 \pm 0.0000\)
& \(0.6170 \pm 0.0141\)
& \(4.97 \pm 0.34\%\)
& \(0.9849 \pm 0.0013\) \\
Feature-visible OOD
& \(5.6563 \pm 0.0098\)
& \(33.2810 \pm 0.8751\)
& \(4.4904 \pm 0.0413\)
& \(4.9156 \pm 0.0602\)
& \(72.20 \pm 2.47\%\)
& \(2.5061 \pm 0.1551\) \\
Feature-collapsed OOD
& \(4.7013 \pm 0.0423\)
& \(3.9992 \pm 0.0617\)
& \(2.2472 \pm 0.0614\)
& \(1.5441 \pm 0.0896\)
& \(17.70 \pm 1.56\%\)
& \(1.0448 \pm 0.0069\) \\
\bottomrule
\end{tabular}
}

\medskip
(b) Predictive performance

\begin{tabular}{llccc}
\toprule
Condition & Method & Coverage & Width & Interval score \\
\midrule
\multirow{3}{*}{ID reference}
& LACP
& \(0.8984 \pm 0.0016\)
& \(2.1772 \pm 0.0403\)
& \(2.7081 \pm 0.0591\) \\
& Unweighted Laplace
& \(0.8996 \pm 0.0015\)
& \(2.0512 \pm 0.0437\)
& \(2.6460 \pm 0.0617\) \\
& CLAPS
& \(0.8989 \pm 0.0014\)
& \(2.1442 \pm 0.0395\)
& \(2.6659 \pm 0.0572\) \\
\midrule
\multirow{3}{*}{Feature-visible OOD}
& LACP
& \(0.1946 \pm 0.0098\)
& \(1.9037 \pm 0.1211\)
& \(57.1267 \pm 1.0449\) \\
& Unweighted Laplace
& \(0.3489 \pm 0.0079\)
& \(3.6711 \pm 0.0918\)
& \(44.7476 \pm 0.7409\) \\
& CLAPS
& \(0.3885 \pm 0.0181\)
& \(4.4043 \pm 0.2273\)
& \(41.7066 \pm 1.4390\) \\
\midrule
\multirow{3}{*}{Feature-collapsed OOD}
& LACP
& \(0.4091 \pm 0.0190\)
& \(1.8861 \pm 0.0618\)
& \(12.3982 \pm 0.9791\) \\
& Unweighted Laplace
& \(0.4084 \pm 0.0184\)
& \(1.8458 \pm 0.0618\)
& \(12.3712 \pm 0.9780\) \\
& CLAPS
& \(0.4286 \pm 0.0188\)
& \(1.9704 \pm 0.0667\)
& \(11.8285 \pm 0.9277\) \\
\bottomrule
\end{tabular}
\end{table}

\subsection{Summary of Sensitivity Analyses}
\label{app:sensitivity_summary}

Across the three sensitivity analyses, marginal coverage remains stable and CLAPS maintains favorable interval scores relative to matched LACP. The improvements are consistently stronger in the weakest-support quintile, where the epistemic contribution is also larger. The sensitivity results therefore connect the predictive changes to the intended support-dependent correction rather than to a uniform change in interval scale.

Prior precision determines the strength of posterior shrinkage. Increasing \(\lambda\) progressively suppresses the epistemic component and moves CLAPS toward aleatoric local scaling. The smooth changes across the fixed grid agree with the reduction analysis, while training-only selection retains a nontrivial correction using only the training data.

Representation dimension controls the feature geometry available to the last-layer posterior. Larger representations support stronger epistemic differentiation and more pronounced weak-support effects, while the direction of the interval-score improvements remains stable across dimensions. The default dimension lies within this stable regime and requires little additional covariance storage or test-time computation.

The variance floor controls the influence of small learned variances on the weighted information matrix. Increasing the floor limits extreme precision weights, while predictive performance and the epistemic profile remain stable. The default floor consequently regularizes the weighted last-layer calculation without materially changing the learned heteroscedastic structure.

The OOD stress test further clarifies the role of the fixed representation. CLAPS produces a strong epistemic response when distribution shift remains visible in learned-feature space, but the correction is substantially weaker when target-relevant changes are compressed into an ID-like representation. The same qualitative limitation appears for the unweighted last-layer variant, showing that the learned feature map constrains both posterior constructions.

Together, the analyses distinguish the roles of the main design components: prior precision controls posterior shrinkage, representation dimension controls the expressiveness of learned support, the variance floor controls numerical weighting, and the fixed representation determines which distribution shifts remain visible to the last-layer correction. The observed changes follow these roles, while the principal marginal and weak-support conclusions remain consistent across the evaluated configurations.

\end{document}